\documentclass{article}

\usepackage{microtype}
\usepackage{graphicx}
\usepackage{subcaption}
\usepackage{booktabs} 

\usepackage{amsmath}
\DeclareMathOperator\sign{sign}

\usepackage{multirow}

\usepackage[draft]{hyperref} 

\usepackage[accepted]{icml2018}

\icmltitlerunning{Learning Confidence for Out-of-Distribution Detection in Neural Networks}

\begin{document}

\twocolumn[
\icmltitle{Learning Confidence for Out-of-Distribution Detection in Neural Networks}



\icmlsetsymbol{equal}{*}

\begin{icmlauthorlist}
\icmlauthor{Terrance DeVries}{g,vi}
\icmlauthor{Graham W. Taylor}{g,vi}
\end{icmlauthorlist}

\icmlaffiliation{g}{School of Engineering, University of Guelph, Guelph, Ontario, Canada}
\icmlaffiliation{vi}{Vector Institute, Toronto, Ontario, Canada}

\icmlcorrespondingauthor{Terrance DeVries}{terrance@uoguelph.ca}
\icmlcorrespondingauthor{Graham W. Taylor}{gwtaylor@uoguelph.ca}

\icmlkeywords{confidence, out-of-distribution, detection, deep learning}

\vskip 0.3in
]



\printAffiliationsAndNotice{}  

\begin{abstract}

Modern neural networks are very powerful predictive models, but they are often incapable of recognizing when their predictions may be wrong. Closely related to this is the task of out-of-distribution detection, where a network must determine whether or not an input is outside of the set on which it is expected to safely perform. To jointly address these issues, we propose a method of learning confidence estimates for neural networks that is simple to implement and produces intuitively interpretable outputs. We demonstrate that on the task of out-of-distribution detection, our technique surpasses recently proposed techniques which construct confidence based on the network's output distribution, without requiring any additional labels or access to out-of-distribution examples. Additionally, we address the problem of calibrating out-of-distribution detectors, where we demonstrate that misclassified in-distribution examples can be used as a proxy for out-of-distribution examples.

\end{abstract}

\section{Introduction}
What do you get if you multiply six by nine? How many roads must a man walk down? What is the meaning of life, the universe, and everything? Some of these questions can be answered confidently, while the answers for others will likely have some amount of associated uncertainty. Knowing the limitations of one's own understanding is important for decision making, as this information can be used to quantify and then minimize the associated potential risk.

Unfortunately, modern neural network classifiers yield very poorly calibrated confidence estimates \citep{guo2017calibration}, and as a result will often produce incorrect predictions with very high predicted class probability. Furthermore, these networks are incapable of identifying inputs that are different from those observed during training \citep{amodei2016concrete}. The result is that these models fail silently, often producing highly confident predictions even when faced with nonsensical inputs \citep{nguyen2015deep,hendrycks2016baseline} or attacked adversarially \citep{szegedy2013intriguing,goodfellow2014explaining}. This undesirable behaviour has brought about concerns for AI Safety \citep{amodei2016concrete}, which have encouraged the development of models that are capable of successfully identifying when they have encountered new situations or inputs, i.e., out-of-distribution examples.

\begin{figure}[t]
\centering
\includegraphics[width=0.5\textwidth, trim={0cm 0cm -1.5cm 0cm}, clip]{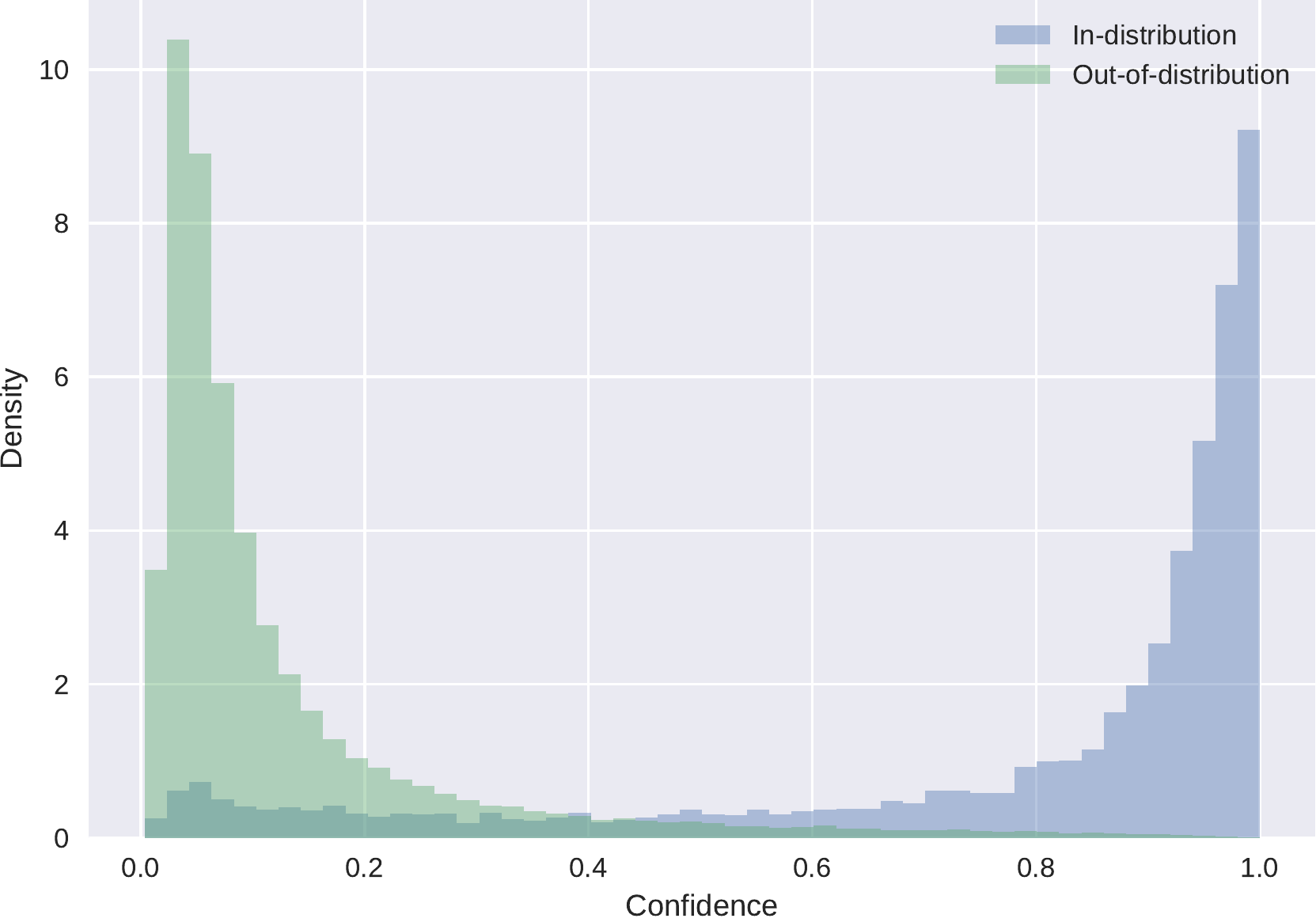}
\caption{Learned confidence estimates can be used to easily separate in- and out-of-distribution examples. Here, the CIFAR-10 test set is used as the in-distribution dataset, and TinyImageNet, LSUN, and iSUN are used as the out-of-distribution datasets. The model is trained using a DenseNet architecture.}
\label{fig:confidence_histogram}
\end{figure}

Recently, several techniques have been proposed to address the problem of out-of-distribution detection. \citet{hendrycks2016baseline} introduce a simple but effective heuristic, which consists of thresholding candidates based on the predicted softmax class probability. \citet{anonymous2018training} propose an improved solution, which involves jointly training a generator and a classifier. The generator produces examples that appear to be at the boundary of the data manifold to serve as out-of-distribution examples, while the classifier is encouraged to assign these uniform class probabilities. Both of these techniques can leverage ODIN \citep{anonymous2018enhancing}, which applies temperature scaling and input preprocessing to further increase the distance between in- and out-of-distribution examples.

In this work, we train neural network classifiers to output confidence estimates for each input, which we then use to differentiate between in and out-of-distribution examples (as shown in Figure~\ref{fig:confidence_histogram}). We demonstrate that this technique improves upon the baseline method \citep{hendrycks2016baseline} and ODIN \citep{anonymous2018enhancing} in almost all test cases, and performs well consistently across several different network architectures. Additionally, our confidence estimation technique produces intuitively interpretable outputs, is simple to implement, and requires very little additional computation over the baseline technique. We also demonstrate that misclassified in-distribution examples can be used as a proxy for out-of-distribution examples when calibrating out-of-distribution detectors, which eliminates the difficulty and expense of collecting or generating out-of-distribution examples.

\begin{figure*}[t]
\centering
\includegraphics[width=0.9\textwidth, trim={0cm 0cm 0cm 0cm}, clip]{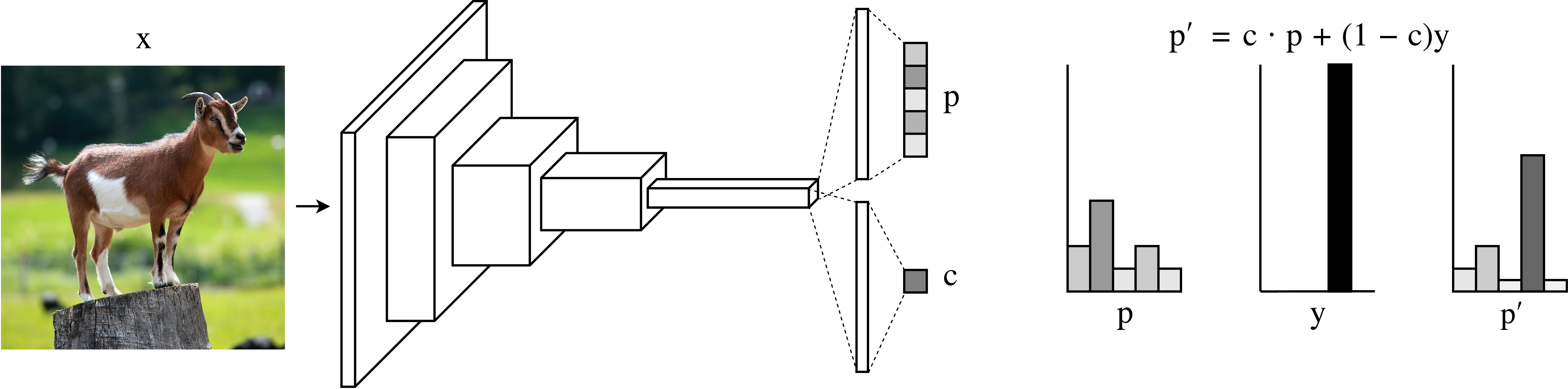}
\caption{Neural network that has been augmented with a confidence estimation branch. The network receives input $x$ and produces softmax prediction probabilities $p$ and a confidence estimate $c$. During training, the predictions are modified according to the confidence of the network such that they are closer to the target probability distribution $y$.}
\label{fig:network_architecture}
\end{figure*}

\section{Confidence Estimation}
The tasks of confidence estimation and out-of-distribution detection are closely related, as would be expected given that we generally have less confidence in our decisions when in foreign situations. As such, a model with well calibrated confidence estimates should also be able to identify out-of-distribution examples. Ideally, we would like to learn a measure of confidence for each input directly, but this proves to be a difficult task as in most machine learning tasks, there are no ground truth labels available for confidence estimation. Instead of learning confidences directly from labels, we introduce an approach in which the neural network is incentivized during training to produce confidence estimates that correctly reflect the model's ability to produce a correct prediction for any given input, in exchange for a reduction in loss.

\subsection{Motivation}
Imagine a test writing scenario. For this particular test, the student is given the option to ask for hints, but for each hint they receive, they also incur some small penalty. In order to optimize their score on the test, a good strategy would be to answer all of the questions that they are confident in without using the hints, and then to ask for hints to the questions that they are uncertain about in order to improve their chances of answering them correctly. At the end of the test, their level of confidence for each question can be approximated by counting the number of hints used. Applying this same strategy to neural networks gives us the ability to learn confidence estimates without the need for any ground truth labels.

\subsection{Learning to Estimate Confidence}
\label{section:learning_confidence}
In order to give neural networks the ability to ask for hints, we first add a confidence estimation branch to any conventional feedforward architecture in parallel with the original class prediction branch, as shown in Figure~\ref{fig:network_architecture}. In practice, the confidence estimation branch is usually added after the penultimate layer of the original network, such that both the confidence branch and the prediction branch receive the same input.

The confidence branch contains one or more fully-connected layers, with the final layer outputting a single scalar between 0 and 1 (parametrized as a sigmoid). This confidence value $c$ represents the network's confidence that it can correctly produce the target output given some input. If the network is confident that it can produce a correct prediction for the given input, it should output $c$ close to 1. Conversely, if the network is not confident that it can produce the correct prediction, then it should output $c$ close to 0.

Prior to normalization, we have a neural network that takes an input $x$ and yields prediction logits and a confidence logit. To the prediction logits we apply the softmax function to obtain class prediction probabilities $p$, while the confidence logit is passed through a sigmoid to obtain the confidence estimate $c$:
\begin{equation}
p,\ c = f(x, \Theta) \quad p_i, c \in [0,1], \sum\limits_{i=1}^M p_i = 1 .\
\end{equation}
In order to give the network ``hints'' during training, the softmax prediction probabilities are adjusted by interpolating between the original predictions and the target probability distribution $y$, where the degree of interpolation is indicated by the network's confidence:
\begin{equation}
p'_{i}  = c \cdot p_{i} + (1 - c) y_{i} .\
\label{eq:pred_interpolation}
\end{equation}
This is demonstrated visually in Figure~\ref{fig:network_architecture}. The task loss is now calculated as usual, using the modified prediction probabilities. For our classification experiments, we use negative log likelihood, but this formulation should generalize to most other loss functions:
\begin{equation}
\mathcal{L}_{t} = -\sum\limits_{i=1}^M \log(p'_{i}) y_{i} .\
\end{equation}
To prevent the network from minimizing the task loss by always choosing $c = 0$
and receiving the entire ground truth, we add a log penalty to the loss function, which we call the confidence loss. This can be interpreted as a binary cross-entropy loss, where the target value is always 1 (i.e.,~we want the network to always be very confident):
\begin{equation}
\mathcal{L}_{c} = -\log(c) .\
\end{equation}
The final loss is simply the sum of the task loss and the confidence loss. The confidence loss is weighted by a hyperparameter $\lambda$, which balances the task loss and the confidence loss:
\begin{equation}
\mathcal{L} = \mathcal{L}_{t} + \lambda\mathcal{L}_{c} .\
\end{equation}
We can now investigate how the confidence score impacts the dynamics of the loss function. In cases where $c \rightarrow 1$ (i.e.,~the network is very confident), we see that $p' \rightarrow p$ and the confidence loss goes to 0. This is equivalent to training a standard network without a confidence branch. In the case where $c \rightarrow 0$ (i.e.~the network is not very confident), we see that $p' \rightarrow y$, so the network receives the correct label. In this scenario the task loss will go to 0, but the confidence loss becomes very large. Finally, if $c$ is some value between 0 and 1, then $p'$ will be pushed closer to the target probabilities, resulting in a reduction in the task loss at the cost of an increase in the confidence loss. This interaction produces an interesting optimization problem, wherein the network can reduce its overall loss if it can successfully predict which inputs it is likely to classify incorrectly.

\subsection{Implementation Details}
Though the approach described in Section~\ref{section:learning_confidence} is simple, it requires several optimizations to make it robust across datasets and architectures. These include methods for automatically selecting the best value for $\lambda$ during training, combating excessive regularization that reduces classification accuracy, and retaining misclassified examples throughout training.

\subsubsection{Budget Parameter}
The first challenge we encounter upon na\"ive application of our method is that as training progresses, $c$ often converges to unity for all samples, such that learned confidence estimates eventually lose their utility. To ensure that the confidence estimates retain meaning throughout training (i.e.~$c\rightarrow1$ for correctly classified samples and $c\rightarrow0$ for incorrectly classified examples), we introduce a budget hyperparameter $\beta$, which represents the amount of confidence penalty that the network is allowed to incur. As training progresses, we adjust the confidence loss weighting $\lambda$ after each weight update such that the confidence loss tends towards $\beta$: if $\mathcal{L}_{c} > \beta$ then increase $\lambda$ (i.e.,~make it more expensive to ask for hints), and if $\mathcal{L}_{c} < \beta$ then decrease $\lambda$ (i.e.,~make it more affordable to ask for hints). We observed that the selection of the budget parameter does not significantly affect the performance of the model on out-of-distribution detection tasks for reasonable values of $\beta$ (e.g., between 0.1 and 1.0).

\subsubsection{Combating Excessive Regularization}
In our experiments, we find that the confidence learning mechanism acts as a very strong regularizer. This can be a desirable property in some cases, as the network will automatically ignore outliers and noisy regions in the dataset. However, we can also encounter the scenario where the model decides to lazily opt for free labels instead of learning complex decision boundaries. While this solution is still valid in our construction of the problem, we want to
encourage the model to take risks and learn from its mistakes, analogous to exploration in reinforcement learning. We find that giving hints with 50\% probability addresses this issue, as the gradients from low confidence examples now have a chance of backpropagating unhindered and updating the decision boundary, but we still learn useful confidence estimates. One way we can implement this in practice is by applying Equation~\ref{eq:pred_interpolation} to only half of the batch at each iteration.

\begin{figure*}[!htbp]
\centering
\begin{minipage}{0.22\textwidth}
 \includegraphics[width=\linewidth, trim={1.0cm 0.75cm 1.5cm 0cm}, clip]{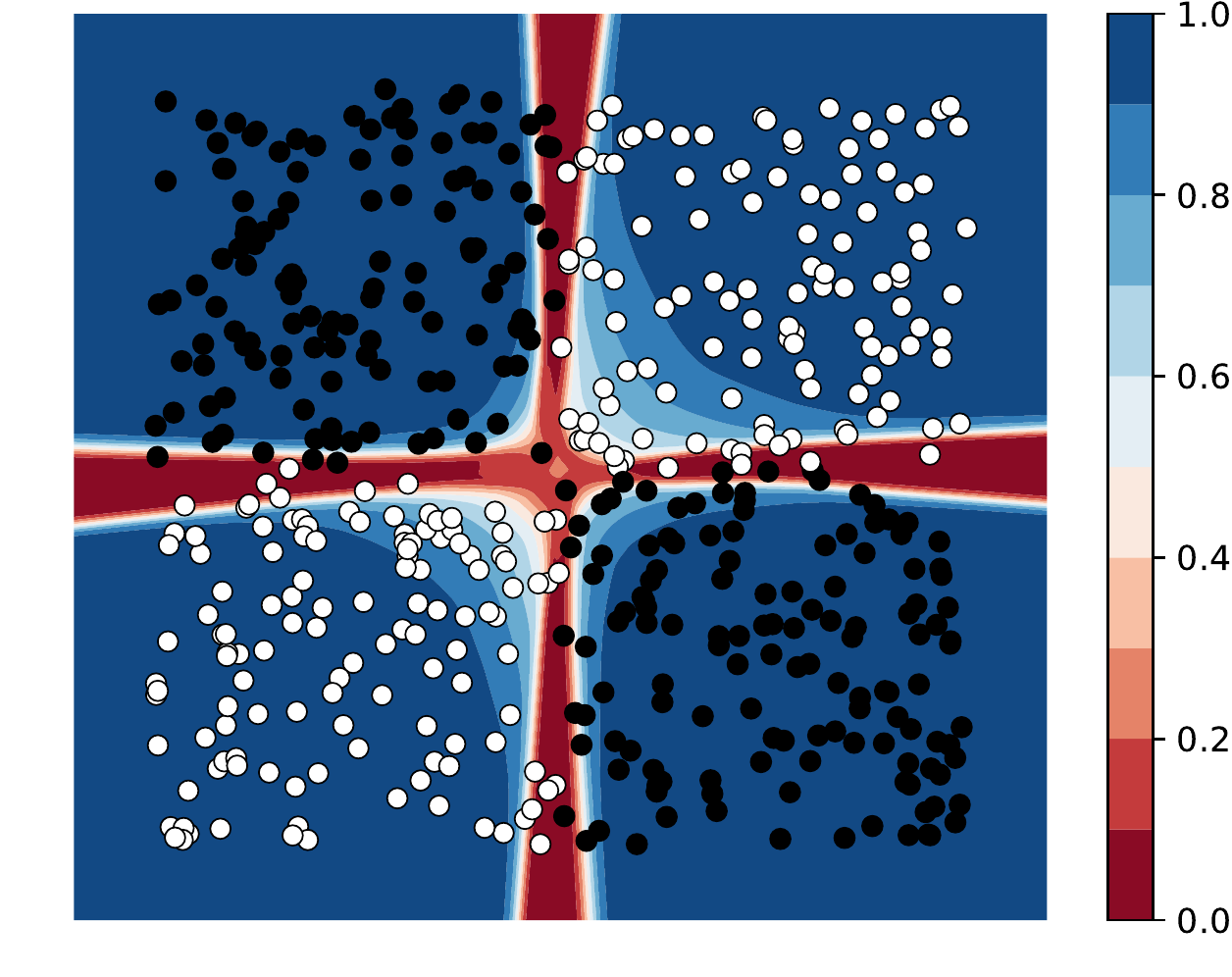}
 \subcaption{Noise = 0\%}
\end{minipage}
\begin{minipage}{0.22\textwidth}
 \includegraphics[width=\linewidth, trim={0.75cm 0.75cm 1.5cm 0cm}, clip]{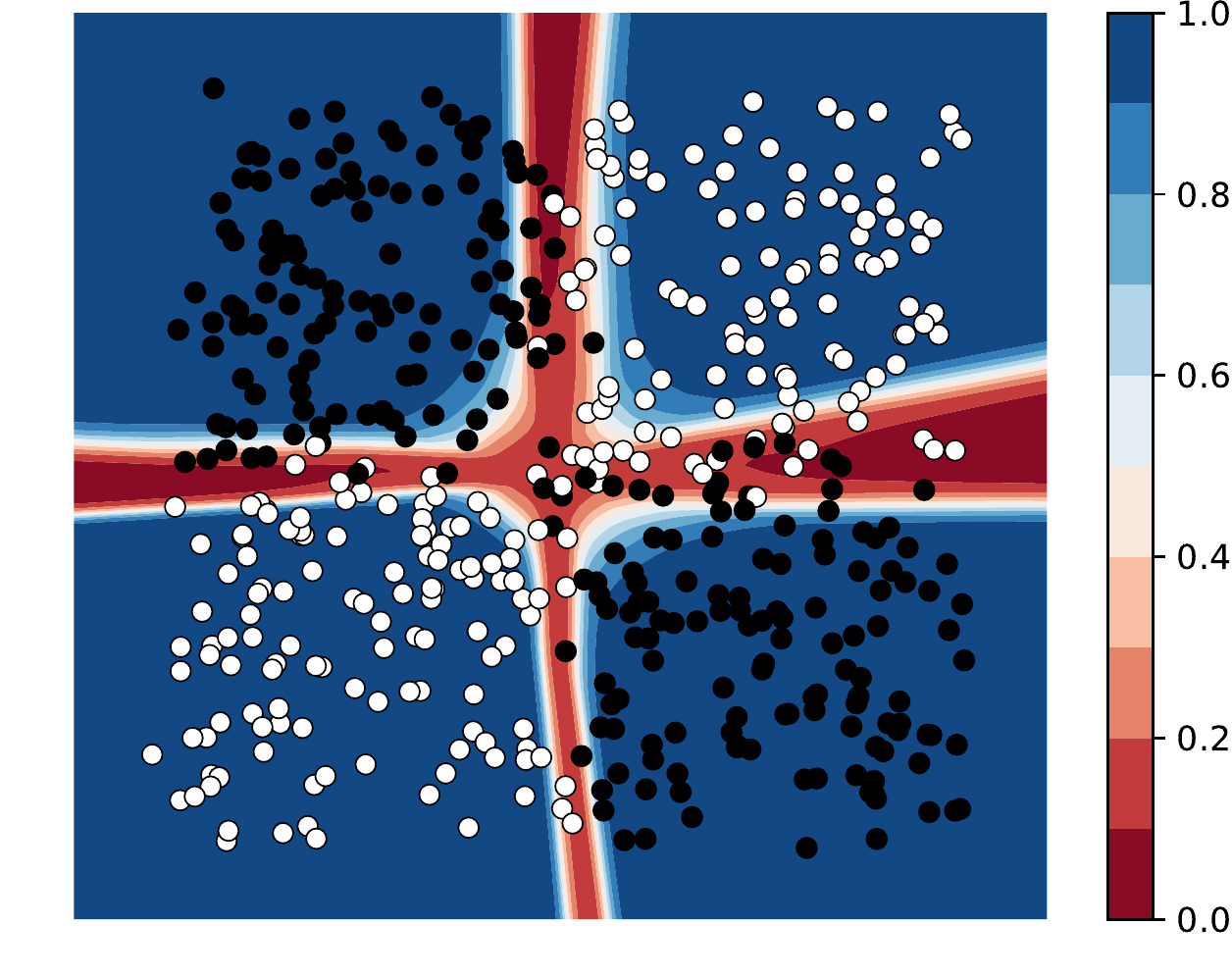}
 \subcaption{Noise = 10\%}
\end{minipage}
\begin{minipage}{0.22\textwidth}
 \includegraphics[width=\linewidth, trim={0.75cm 0.75cm 1.5cm 0cm}, clip]{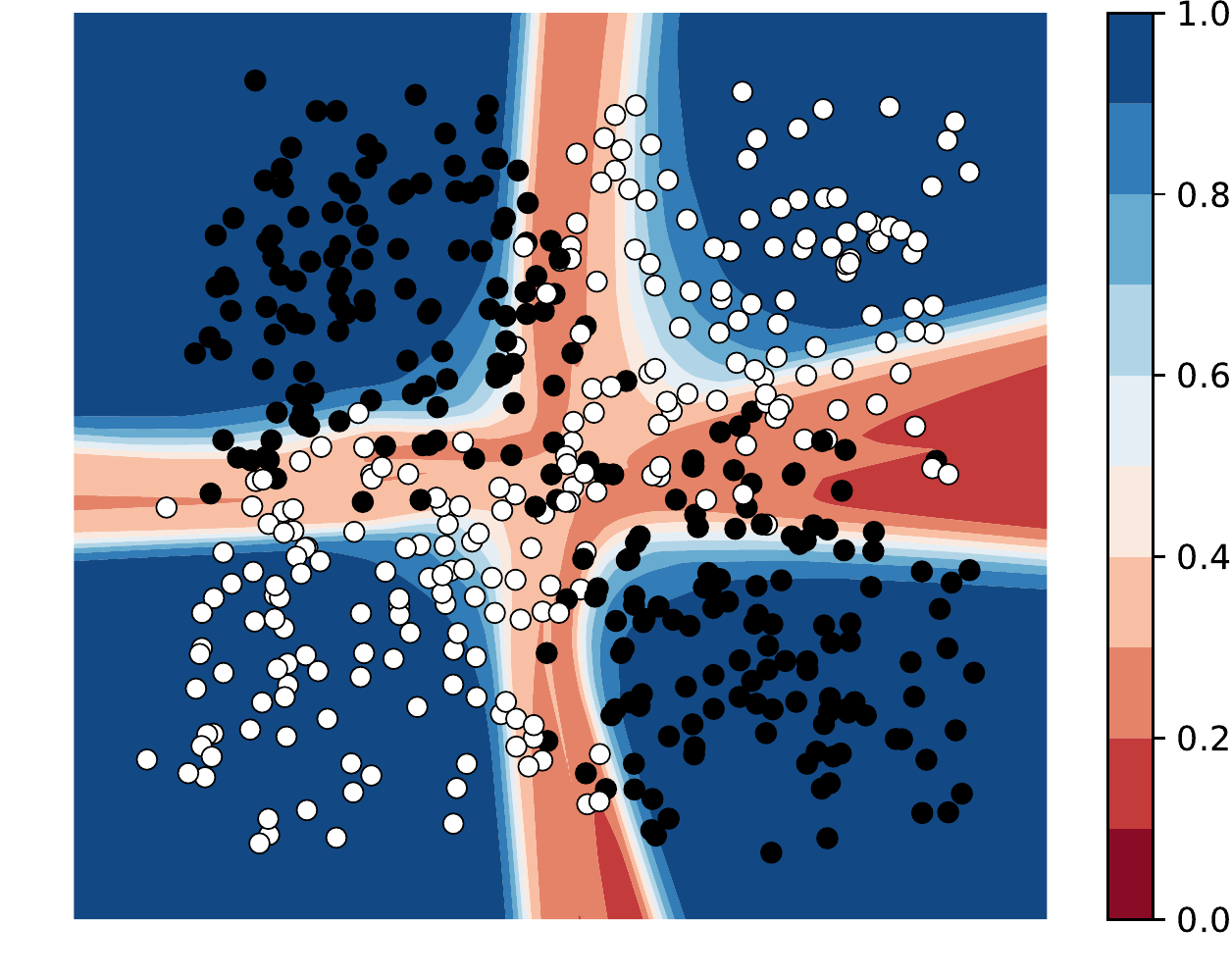}
 \subcaption{Noise = 20\%}
\end{minipage}
\begin{minipage}{0.247\textwidth}
 \includegraphics[width=\linewidth, trim={0.75cm 0.5cm 0cm 0cm}, clip]{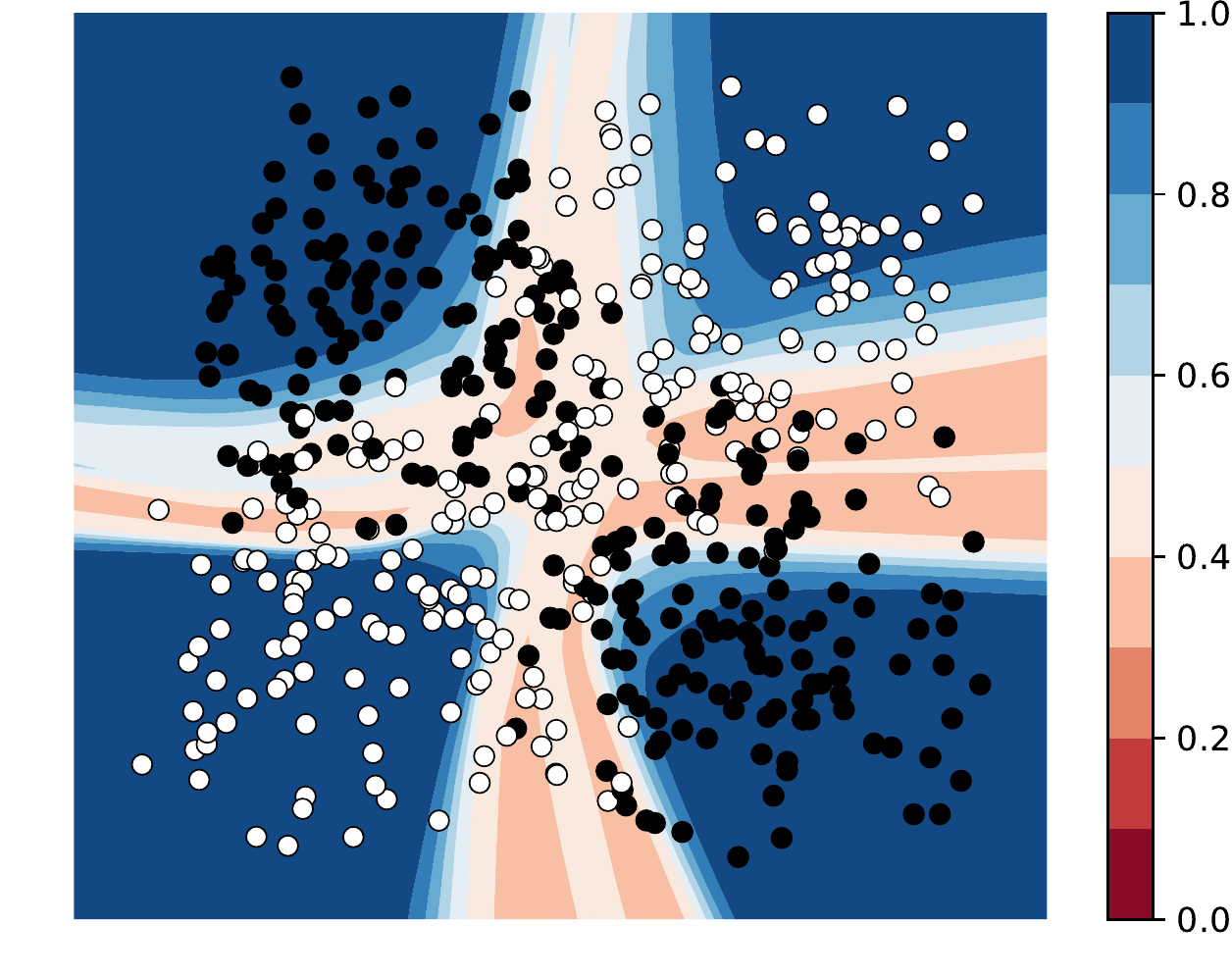}
 \subcaption{Noise = 30\%}
\end{minipage}\\
\caption{Confidence predictions on XOR dataset with varying levels of noise and a fixed budget of $\beta$ = 0.3. Blue indicates high confidence, and red indicates low confidence. Best viewed in colour.}
\label{fig:2D_plots}
\end{figure*}

\subsubsection{Retaining Misclassified Examples}
When training high-capacity neural network architectures on small datasets, we commonly observe excessive overfitting on the training set, where the model learns to correctly classify every training sample presented to it. This behaviour is detrimental to training confidence estimators, since misclassified examples are necessary in order for the model to learn the concept of confidence. Without misclassified examples available, the model begins assigning low confidence to samples that it should be able to classify easily, which quickly degrades the quality of the confidence estimation. To prevent this situation from occurring, we find that aggressive data augmentation can be used to create difficult examples and prevent overfitting. In our experiments, we apply standard random cropping and flipping data augmentations, as well as Cutout \citep{devries2017improved}, which randomly masks regions of the input images. Other regularizers such as early stopping may also be effective in mitigating this effect.

\section{Out-of-Distribution Detection}
Once we have trained a model, we can use the learned confidence estimates to perform out-of-distribution detection. Specifically, we evaluate the function:

\begin{equation}
  g(x;\delta)=\begin{cases}
    1 & \text{if $c \leq \delta$}\\
    0 & \text{if $ c > \delta$}
  \end{cases}
\end{equation}

Where input $x$ is marked as out-of-distribution if the confidence estimate $c$ is less than or equal to some detection threshold $\delta$. Selecting a good threshold value is crucial for the success of the out-of-distribution detector, as the final quality of the detector can sometimes vary widely even for small changes in $\delta$. We investigate the impact of selecting $\delta$ in Section~\ref{sec:selecting_detection_threshold}.

\subsection{Input Preprocessing}
\label{section: confidence_scaling}
In order to further separate the in- and out-of-distribution examples, we apply an input preprocessing technique based on an idea proposed by \citet{anonymous2018enhancing}. This technique was originally inspired by the fast gradient sign method (FGSM) of generating adversarial examples \citep{goodfellow2014explaining}. The goal of FGSM is to add small perturbations to an image in order to decrease the softmax prediction probability of the target class, ultimately in an attempt to make a model misclassify that image. In essence, FGSM nudges the image away from the correct class. \citet{anonymous2018enhancing} use this technique in the opposite manner, that is, to push images closer to their \emph{predicted} class. They find that in-distribution images are drawn closer to the predicted class after this kind of input preprocessing when compared to out-of-distribution examples, which makes the two distributions easier to separate. We apply input preprocessing in a similar fashion; however, we perturb images such that they are pushed towards being more confident ($c \rightarrow 1)$. To calculate the necessary perturbation, we simply backpropagate the gradients of the confidence loss with respect to the inputs:

\begin{equation}
\tilde{x} = x - \epsilon \sign (\nabla_{x}\mathcal{L}_{c}) \
\label{eq:input_preprocessing}
\end{equation}

Where $\epsilon$ represents the magnitude of the noise perturbation that is added to each image. Similar to \citet{anonymous2018enhancing}, we observe that in-distribution examples increase in confidence more than out-of-distribution examples using this procedure, resulting in an easier separation of the two distributions.

\section{Experiments}
To better understand the behaviour of the confidence estimator, we first experiment with a toy 2D dataset. We then evaluate the effectiveness of our confidence score as a means of separating in- and out-of-distribution examples by replicating several experiments originally conducted by \citet{hendrycks2016baseline} and \citet{anonymous2018enhancing}.

\subsection{Visualizing Learned Confidence Estimates}

We begin by visualizing the behaviour of the learned confidence estimates on a simple 2D XOR dataset. For these tests we generate several datasets consisting of 500 training samples, each with progressively increasing noise levels. Our model architecture is an MLP with 3 layers with 100 hidden units each, followed by parallel classification and confidence branches, each with a 100 hidden unit layer. We train the model using SGD with batches of 10 samples for 30 epochs. The budget $\beta$ is set to 0.3 for all tests.

Our first observation from this exploration (Figure~\ref{fig:2D_plots}) is that the confidence branch appears to be making reasonable estimates; it outputs low confidence in noisy regions that contain datapoints from both classes (i.e.~along true class boundaries), and high confidence when in regions that only contain a single class. We also observe that as the amount of noise in the dataset increases for a fixed budget, the lower bound on the confidence estimate increases accordingly. This indicates that a larger budget may be required for noisier datasets in order to maintain a confidence close to 0 for out-of-distribution and misclassified examples. Test results for a broader range of parameters can be found in Figure \ref{fig:checkerboard_grid} of the Appendix.

\subsection{Out-of-Distribution Detection}
To evaluate how suitable the learned confidence estimates are for separating in- and out-of-distribution examples, we replicate experiments introduced by \citet{hendrycks2016baseline} and \citet{anonymous2018enhancing} for evaluating out-of-distribution detection methods. In these experiments, a neural network is first trained on some dataset, which represents the in-distribution examples. Out-of-distribution examples are represented by images from a variety of datasets that contain classes different from those found in the in-distribution dataset. For each sample in the in-distribution test set, and each out-of-distribution example, a confidence score is produced, which will be used to predict which distribution the samples come from. Finally, several different evaluation metrics (defined in Section~\ref{evaluation_metrics}) are used to measure and compare how well different confidence estimation methods can separate the two distributions.

\subsubsection{In-Distribution Dataset}
For our in-distribution examples, we use two common image classification datasets: SVHN and CIFAR-10. For both datasets, we train a model on the training set, and then use the test set for evaluating both classification accuracy and the performance of out-of-distribution detection methods.

\textbf{SVHN:} The Street View Housing Numbers (SVHN) dataset \citep{netzer2011reading} consists of colour images depicting house numbers, which range from 0 to 9. Images have a resolution of $32\times32$. For our tests, we use the official training set split which contains 73,257 images, and the test set split, which has 26,032 images.

\textbf{CIFAR-10:} The CIFAR-10 dataset \citep{krizhevsky2009learning} consists of natural colour images, each of size 32 $\times$ 32 pixels. Each image is classified into 1 of 10 classes, such as dog, cat, automobile, or ship. The training set contains 50,000 images, while the test set contains 10,000 images.

\subsubsection{Out-of-Distribution Datasets}
For our tests, we use the same out-of-distribution datasets used by \citet{anonymous2018enhancing}, most of which are available for download on their GitHub project page\footnote{https://github.com/ShiyuLiang/odin-pytorch}. The datasets are:

\textbf{TinyImageNet:} The TinyImageNet dataset\footnote{https://tiny-imagenet.herokuapp.com/} is a subset of the ImageNet dataset \citep{deng2009imagenet}. The test set for TinyImageNet contains 10,000 images from 200 different classes, and is used to create two out-of-distribution datasets, also containing 10,000 images each. \textit{TinyImageNet\ (crop)} is made by randomly cropping a patch of size 32 $\times$ 32 from each test image, while \textit{TinyImageNet\ (resize)} contains the original images, downsampled to 32 $\times$ 32 pixels.\\
\textbf{LSUN:} The Large-scale Scene UNderstanding dataset (LSUN) \citep{yu2015lsun} has a test set consisting of 10,000 images from 10 different scene classes, such as bedroom, church, kitchen, and tower. Two different out-of-distribution datsets, \textit{LSUN\ (crop)} and \textit{LSUN\ (resize)} are created in a similar fashion to the TinyImageNet datasets, by randomly cropping and downsampling the test set, respectively.\\
\textbf{iSUN:} The iSUN dataset \citep{xu2015turkergaze} is a subset of the SUN dataset, containing 8,925 images. All images in this dataset are used, downsampled to 32 $\times$ 32 pixels.\\
\textbf{Uniform Noise:} The uniform noise dataset is generated by drawing each pixel in a 32 $\times$ 32 RGB image from an i.i.d uniform distribution of the range [0, 1]. The dataset contains 10,000 samples in total.\\
\textbf{Gaussian Noise:} The Gaussian noise dataset is generated by drawing each pixel in a 32 $\times$ 32 RGB image from an i.i.d Gaussian distribution with a mean of 0.5 and variance of 1. The pixel values in each image are clipped to the range [0, 1] in order to keep them in the expected range for images. The dataset contains 10,000 samples in total.\\
\textbf{All Images:} This dataset is a combination of all real image datasets: TinyImageNet (crop), TinyImageNet (resize), LSUN (crop), LSUN (resize), and iSUN. Note that in this case, a single threshold must be used for all datasets, so this scenario best mimics the situation that would be faced by a real world out-of-distribution detector.

\begin{table*}[t]
\caption{Comparison of the baseline method \citep{hendrycks2016baseline} and confidence-based thresholding. All results are averaged over 5 runs. All values are shown in percentages. $\downarrow$ indicates that lower values are better, while $\uparrow$ indicates that higher scores are better.}
\vspace{0.1in}
\small
\centering
\begin{tabular}{lccccccc}
\hline
\begin{tabular}[l]{@{}l@{}}\textbf{Model}\\ In-distribution \\Dataset\end{tabular} & \begin{tabular}[c]{@{}c@{}}\textbf{Out-of-distribution}\\ \textbf{Dataset}\end{tabular} & \multicolumn{1}{c}{\begin{tabular}[c]{@{}c@{}}\textbf{Classification} \\ \textbf{Error} \\ $\downarrow$\end{tabular}} & \multicolumn{1}{c}{\begin{tabular}[c]{@{}c@{}}\textbf{FPR}\\ \textbf{(95\% TPR)}\\ $\downarrow$\end{tabular}} & \multicolumn{1}{c}{\begin{tabular}[c]{@{}c@{}}\textbf{Detection}\\ \textbf{Error}\\ $\downarrow$\end{tabular}} & \multicolumn{1}{c}{\begin{tabular}[c]{@{}c@{}}\textbf{AUROC}\\ \\ $\uparrow$\end{tabular}} & \multicolumn{1}{c}{\begin{tabular}[c]{@{}c@{}}\textbf{AUPR}\\ \textbf{In}\\ $\uparrow$\end{tabular}} & \multicolumn{1}{c}{\begin{tabular}[c]{@{}c@{}}\textbf{AUPR}\\ \textbf{Out}\\ $\uparrow$\end{tabular}} \\
\hline
& & \multicolumn{6}{c}{\textbf{Baseline \cite{hendrycks2016baseline}/Confidence Thresholding}} \\
\cline{3-8}
\multirow{3}{*}{\textbf{DenseNet-BC}} & TinyImageNet (resize) & \multirow{4}{*}{2.89/\textbf{2.77}} & 7.2/\textbf{1.5} & 5.3/\textbf{2.8} & 98.4/\textbf{99.5} & 99.4/\textbf{99.8} & 95.6/\textbf{98.7} \\
\multirow{3}{*}{SVHN} & LSUN (resize) & & 6.0/\textbf{1.0} & 4.9/\textbf{2.3} & 98.6/\textbf{99.7} & 99.5/\textbf{99.9} & 96.0/\textbf{99.0} \\
 & iSUN & & 6.0/\textbf{0.9} & 4.9/\textbf{2.3} & 98.6/\textbf{99.7} & 99.5/\textbf{99.9} & 95.7/\textbf{98.8} \\
 & All Images & & 12.2/\textbf{4.2} & 7.2/\textbf{4.5} & 97.3/\textbf{98.9} & 95.1/\textbf{97.4} & 98.4/\textbf{99.4} \\
\hline

\multirow{3}{*}{\textbf{WRN-16-8}} & TinyImageNet (resize) & \multirow{4}{*}{2.77/\textbf{2.66}} & 10.6/\textbf{1.5} & 6.1/\textbf{2.7} & 97.8/\textbf{99.6} & 99.2/\textbf{99.8} & 93.6/\textbf{99.2} \\
\multirow{3}{*}{SVHN} & LSUN (resize) & & 9.5/\textbf{0.6} & 5.8/\textbf{1.8} & 98.0/\textbf{99.8} & 99.3/\textbf{99.9} & 94.0/\textbf{99.5} \\
 & iSUN & & 9.6/\textbf{0.8} & 5.9/\textbf{2.1} & 98.0/\textbf{99.8} & 99.3/\textbf{99.9} & 93.4/\textbf{99.4} \\
 & All Images & & 15.7/\textbf{5.3} & 7.9/\textbf{5.0} & 96.7/\textbf{98.7} & 94.1/\textbf{96.8} & 97.9/\textbf{99.4} \\
\hline

\multirow{3}{*}{\textbf{VGG13}} & TinyImageNet (resize) & \multirow{4}{*}{3.05/\textbf{2.98}} & 11.4/\textbf{1.8} & 6.2/\textbf{3.1} & 97.8/\textbf{99.6} & 99.2/\textbf{99.8} & 93.7/\textbf{99.1} \\
\multirow{3}{*}{SVHN} & LSUN (resize) & & 9.4/\textbf{0.8} & 5.7/\textbf{2.0} & 98.1/\textbf{99.8} & 99.3/\textbf{99.9} & 94.3/\textbf{99.6} \\
 & iSUN & & 10.0/\textbf{1.0} & 6.0/\textbf{2.2} & 98.0/\textbf{99.8} & 99.3/\textbf{99.9} & 93.7/\textbf{99.5} \\
 & All Images & & 14.2/\textbf{4.3} & 7.1/\textbf{4.6} & 97.3/\textbf{99.2} & 95.9/\textbf{98.5} & 98.2/\textbf{99.6} \\
\hline

\multirow{3}{*}{\textbf{DenseNet-BC}} & TinyImageNet (resize) & \multirow{4}{*}{\textbf{4.17}/4.39} & 44.9/\textbf{33.8} & 12.8/\textbf{12.3} & 93.2/\textbf{94.2} & 94.6/\textbf{95.0} & 91.2/\textbf{93.0} \\
\multirow{3}{*}{CIFAR-10} & LSUN (resize) & & 38.6/\textbf{30.7} & 10.8/\textbf{10.3} & 94.6/\textbf{95.4} & 95.9/\textbf{96.4} & 92.8/\textbf{93.9} \\
 & iSUN & & 41.4/\textbf{31.6} & 11.6/\textbf{11.0} & 94.1/\textbf{95.0} & 95.8/\textbf{96.3} & 91.3/\textbf{93.0} \\
 & All Images & & 40.9/\textbf{28.9} & 11.6/\textbf{10.9} & 94.1/\textbf{95.3} & 87.6/\textbf{88.1} & 98.3/\textbf{98.7} \\
\hline

\multirow{3}{*}{\textbf{WRN-28-10}} & TinyImageNet (resize) & \multirow{4}{*}{\textbf{3.25}/3.46} & 41.0/\textbf{26.6} & 14.3/\textbf{11.6} & 91.0/\textbf{94.5} & 88.9/\textbf{94.1} & 90.5/\textbf{94.0} \\
\multirow{3}{*}{CIFAR-10} & LSUN (resize) & & 34.7/\textbf{24.0} & 11.7/\textbf{9.1} & 93.7/\textbf{96.0} & 93.4/\textbf{96.6} & 92.7/\textbf{94.5} \\
 & iSUN &  & 36.7/\textbf{24.9} & 12.6/\textbf{9.8} & 92.8/\textbf{95.7} & 92.6/\textbf{96.5} & 91.1/\textbf{94.0} \\
 & All Images & & 36.1/\textbf{23.3} & 12.4/\textbf{9.7} & 92.9/\textbf{95.7} & 73.3/\textbf{86.7} & 98.1/\textbf{98.8} \\
\hline

\multirow{3}{*}{\textbf{VGG13}} & TinyImageNet (resize) & \multirow{4}{*}{\textbf{5.28}/5.44} & 43.8/\textbf{18.4} & 12.0/\textbf{9.4} & 93.5/\textbf{97.0} & 94.6/\textbf{97.3} & 91.7/\textbf{96.9} \\
\multirow{3}{*}{CIFAR-10} & LSUN (resize) & & 41.9/\textbf{16.4} & 11.5/\textbf{8.3} & 94.0/\textbf{97.5} & 95.1/\textbf{97.8} & 92.2/\textbf{97.2} \\
 & iSUN & & 41.2/\textbf{16.3} & 11.4/\textbf{8.5} & 94.0/\textbf{97.5} & 95.5/\textbf{98.0} & 91.5/\textbf{96.9} \\
 & All Images & & 41.6/\textbf{19.2} & 11.7/\textbf{9.1} & 93.9/\textbf{97.1} & 85.5/\textbf{92.0} & 98.2/\textbf{99.3} \\
\hline
\end{tabular}
\label{table:baseline_confidence}
\end{table*}

\subsubsection{Evaluation Metrics}
\label{evaluation_metrics}

We measure the quality of out-of-distribution detection using the established metrics for this task \citep{hendrycks2016baseline,anonymous2018enhancing}.

\textbf{FPR at 95\% TPR:} Measures the false positive rate (FPR) when the true positive rate (TPR) is equal to $95\%$. Let TP, FP, TN, and FN represent true positives, false positives, true negatives, and false negatives, respectively. The false positive rate is calculated as FPR=FP/(FP+TN), while true positive rate is calculated as TPR=TP/(TP+FN). \\
\textbf{Detection Error:} Measures the minimum possible misclassification probability over all possible thresholds $\delta$ when separating in- and out-of-distribution examples, as defined by $\min_{\delta}\left\{0.5P_{in}(f(x)\leq \delta) + 0.5P_{out}(f(x)> \delta) \right\}$. Here, we equally weight $P_{in}$ and $P_{out}$ as if they have the same probability of appearing in the test set.\\
\textbf{AUROC:} Measures the Area Under the Receiver Operating Characteristic curve. The Receiver Operating Characteristic (ROC) curve plots the relationship between TPR and FPR. The area under the ROC curve can be interpreted as the probability that a positive example (in-distribution) will have a higher detection score than a negative example (out-of-distribution).\\
\textbf{AUPR:} Measures the Area Under the Precision-Recall (PR) curve. The PR curve is made by plotting $\text{precision}=\text{TP}/(\text{TP}+\text{FP})$ versus $\text{recall}=\text{TP}/(\text{TP}+\text{FN})$. In our tests, AUPR-In indicates that in-distribution examples are used as the positive class, while AUPR-Out indicates that out-of-distribution examples are used as the positive class.

\begin{table*}[t]
\centering
\caption{Comparison of ODIN \citep{anonymous2018enhancing} and confidence with input preprocessing when the All Images dataset is used as out-of-distribution. All results are averaged over 5 runs. All values are shown in percentages. $\downarrow$ indicates that lower values are better, while $\uparrow$ indicates that higher scores are better.}
\vspace{0.1in}
\small
\begin{tabular}{lcccccc}
\hline
 & \textbf{\begin{tabular}[c]{@{}c@{}}In-distribution\\ dataset\end{tabular}} & \textbf{\begin{tabular}[c]{@{}c@{}}FPR\\ (95\% TPR)\\ $\downarrow$\end{tabular}} & \textbf{\begin{tabular}[c]{@{}c@{}}Detection\\ Error\\ $\downarrow$\end{tabular}} & \textbf{\begin{tabular}[c]{@{}c@{}}AUROC\\ \\  $\uparrow$\end{tabular}} & \textbf{\begin{tabular}[c]{@{}c@{}}AUPR\\ In\\ $\uparrow$\end{tabular}} & \textbf{\begin{tabular}[c]{@{}c@{}}AUPR\\ Out\\  $\uparrow$\end{tabular}} \\
 \hline
 &  & \multicolumn{5}{c}{\textbf{ODIN \citep{anonymous2018enhancing}/Confidence + Input Preprocessing}} \\
 \cline{3-7}
\multirow{2}{*}{\textbf{DenseNet-BC}}  & SVHN & 8.6/\textbf{4.2} & 6.8/\textbf{4.5} & 97.2/\textbf{98.9} & 92.5/\textbf{97.5} & 98.6/\textbf{99.4} \\
& CIFAR-10 & \textbf{7.8}/16.2 & \textbf{6.0}/8.6 & \textbf{98.4}/97.0 & \textbf{95.3}/91.4 & \textbf{99.6}/99.2 \\
 \hline
{\multirow{2}{*}{\textbf{WideResNet}}} & SVHN & 13.4/\textbf{5.2} & 8.4/\textbf{5.0} & 96.5/\textbf{98.7} & 92.2/\textbf{97.1} & 97.9/\textbf{99.3} \\
\multicolumn{1}{c}{} & CIFAR-10 & 25.0/\textbf{18.9} & 12.0/\textbf{9.4} & 93.4/\textbf{96.2} & 71.6/\textbf{87.2} & 98.3/\textbf{99.0} \\
\hline
\multirow{2}{*}{\textbf{VGG13}} & SVHN & 7.3/\textbf{4.1} & 6.0/\textbf{4.5} & 98.2/\textbf{99.2} & 96.8/\textbf{98.6} & 98.9/\textbf{99.5} \\
& CIFAR-10 & 20.2/\textbf{11.2} & 10.2/\textbf{6.9} & 95.8/\textbf{98.0} & 85.9/\textbf{94.5} & 98.9/\textbf{99.5} \\
 \hline
\end{tabular}
\label{odin_confidence_scaling_table}
\end{table*}

\subsubsection{Model Training}
We evaluate our confidence estimation technique by applying it to several different neural network architectures: DenseNet \citep{huang2017densely}, WideResNet, \citep{Zagoruyko2016WRN}, and VGGNet \citep{simonyan2014very}. Following \citet{anonymous2018enhancing}, we train a DenseNet with depth $L=100$ and growth rate $k=12$ (hereafter referred to as DenseNet-BC), as well as a WideResNet. We use a depth of 16 and a widening factor of 8 for the SVHN dataset, and a depth of 28 and widening factor of 10 for CIFAR-10 (referred to as WRN-16-8 and WRN-28-10 respectively). We also train a VGG13 model to evaluate performance on network architectures without skip connections.\looseness=-1

All models are trained using stochastic gradient descent, with Nesterov momentum of 0.9. We also apply standard data augmentation (cropping and flipping) and Cutout. Following \citet{huang2017densely}, DenseNet-BC is trained for 300 epochs with batches of 64 images, and a weight decay of 1e-4. The learning rate is initialized to 0.1 at the beginning of training and is reduced by a factor of 10$\times$ after the 150th and 225th epochs. We use the settings of \citet{Zagoruyko2016WRN} for both WideResNet and VGG13, training for 200 epochs with batches of size 128, and a weight decay of 5e-4. The learning rate is initialized to 0.1 at the beginning of training, and reduced by a factor of 5$\times$ after the 60\textsuperscript{th}, 120\textsuperscript{th}, and 160\textsuperscript{th} epochs.

We train two sets of models, each with 5 runs per model. The first set of models are trained without any confidence branch, and are used to evaluate the baseline method \citep{hendrycks2016baseline} and ODIN \citep{anonymous2018enhancing}. The second set of models is trained with a confidence estimation branch, and are used to evaluate confidence thresholding with and without input preprocessing. A budget value of $\beta=0.3$ is used for all confidence estimation models. We select $\beta$ such that the optimal detection threshold of our models is around 0.5, which produces confidence estimates that are intuitive to a user of such a system. The test error rates of the trained models are shown in Table~\ref{table:baseline_confidence}. In general, we see that from the perspective of test accuracy, models trained with confidence estimation perform similarly to the baseline. This suggests that irrespective of the \emph{use} of confidence at inference time, training with a confidence branch does not degrade nor significantly improve the performance of a model.\looseness=-1

\subsubsection{Comparison with Baseline}
\label{section:baseline_comparison}
The results of the out-of-distribution detection tests are shown in Table~\ref{table:baseline_confidence} (due to space constrains we only show a portion of the out-of-distribution datasets; complete results can be found in the Appendix). For reference, we compare our technique with the performance of the baseline \citep{hendrycks2016baseline}. We find that thresholding on the learned confidence estimate rather than the softmax prediction probability yields better separation of in- and out-of-distribution examples for almost all network architectures and out-of-distribution datasets.

\subsubsection{Comparison with ODIN}
\label{section:ODIN_comparison}
Recently, \citet{anonymous2018enhancing} introduced the Out-of-Distribution Detector for Neural Networks (ODIN), which improves upon the baseline technique by using temperature scaling and input preprocessing. For fair comparison, we adapt input preprocessing to work with confidence estimates, as described in Section~\ref{section: confidence_scaling}. Following the recommendations of \citet{anonymous2018enhancing}, we use $T=1000$ for all temperature scaling. We also perform a grid search over $\epsilon$ for input preprocessing by selecting the value that achieves the best detection error on a holdout set of 1000 images from each of the out-of-distribution datasets. For this comparison, we only evaluate on the All Images dataset, which provides a good summary of performance across all out-of-distribution datasets. We find that confidence estimates with input preprocessing outperform ODIN in all test settings, with the exception of DenseNet on CIFAR-10 (shown in Table~\ref{odin_confidence_scaling_table}).

\subsubsection{Selecting A Detection Threshold}{}
\label{sec:selecting_detection_threshold}
One challenging aspect of out-of-distribution detection that is not often addressed in the literature is the problem of selecting the detection threshold $\delta$. There are two common scenarios that we can expect to encounter in the problem of out-of-distribution detection, which we address below.

The first situation is the best-case scenario, where we have access to a small set of out-of-distribution examples that can be used for calibration. In this setting, we can calibrate $\delta$ by evaluating the detection error at many different thresholds, and then selecting the one which produces the lowest detection error. Our tests in Sections~\ref{section:baseline_comparison} and~\ref{section:ODIN_comparison} fit into this case. The second situation we may encounter is one in which we do not have access to any out-of-distribution examples, which increases the difficulty of selecting a good value for $\delta$. However, we find that misclassified in-distribution examples from a holdout set (such as a validation set) can serve as a conservative proxy for out-of-distribution examples when calibrating the detection threshold.

\begin{figure*}[htbp]
\centering
\begin{minipage}{0.30\textwidth}
 \includegraphics[width=\linewidth, trim={0cm 0cm 0cm 0cm}, clip]{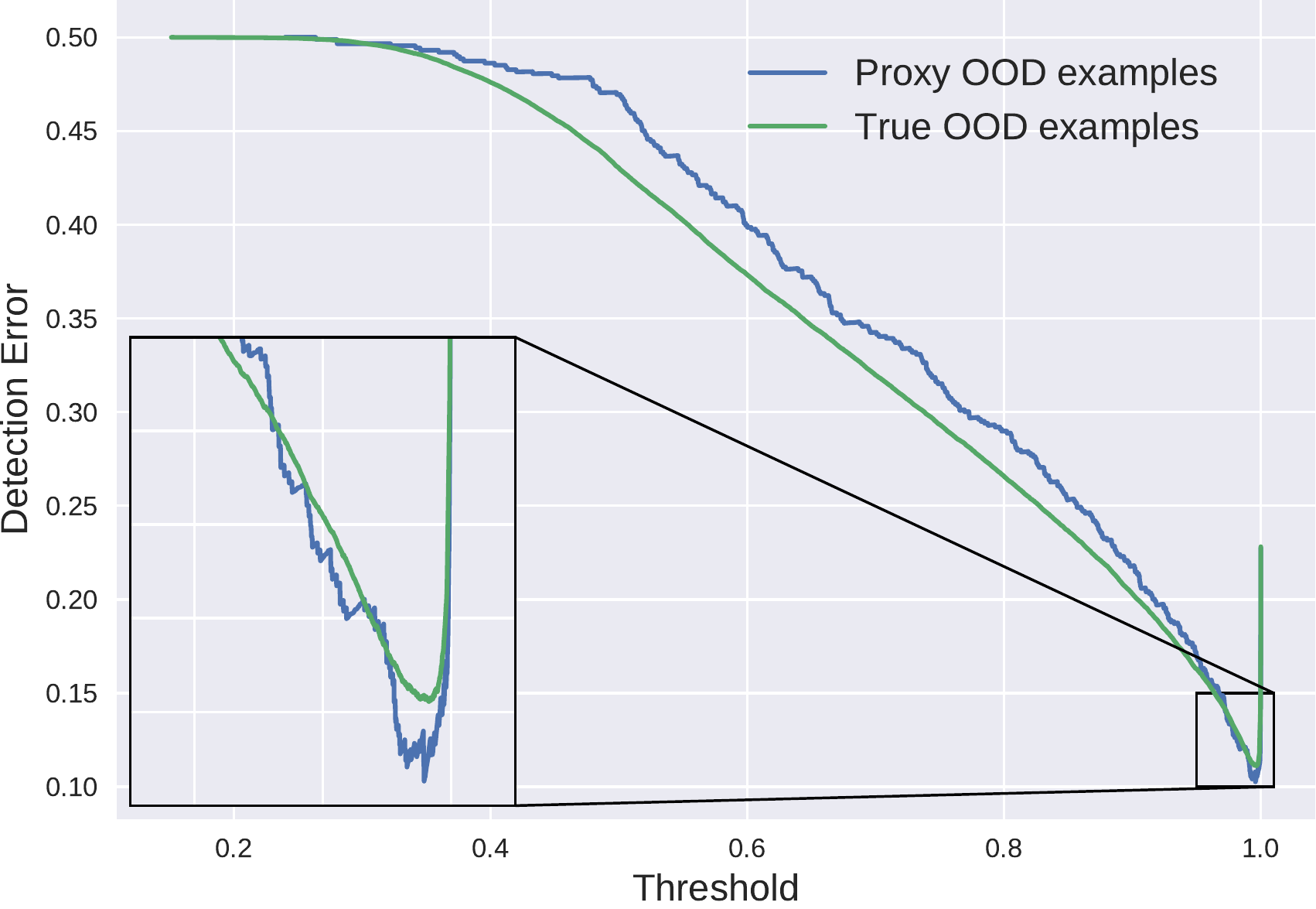}
 \subcaption{DenseNet (baseline)}
\end{minipage}
\begin{minipage}{0.30\textwidth}
 \includegraphics[width=\linewidth, trim={0cm 0cm 0cm 0cm}, clip]{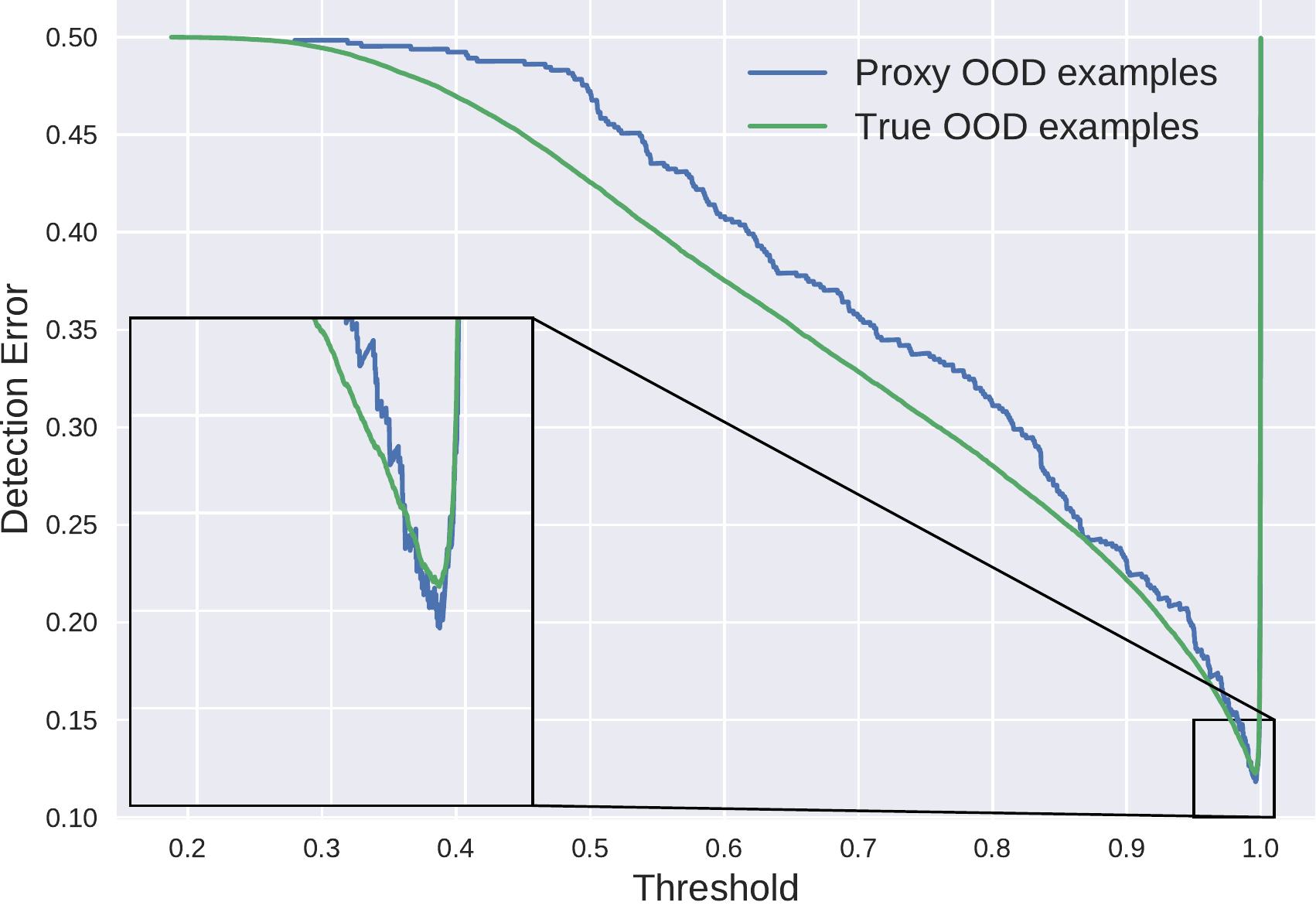}
 \subcaption{WideResNet (baseline)}
\end{minipage}
\begin{minipage}{0.30\textwidth}
 \includegraphics[width=\linewidth, trim={0cm 0cm 0cm 0cm}, clip]{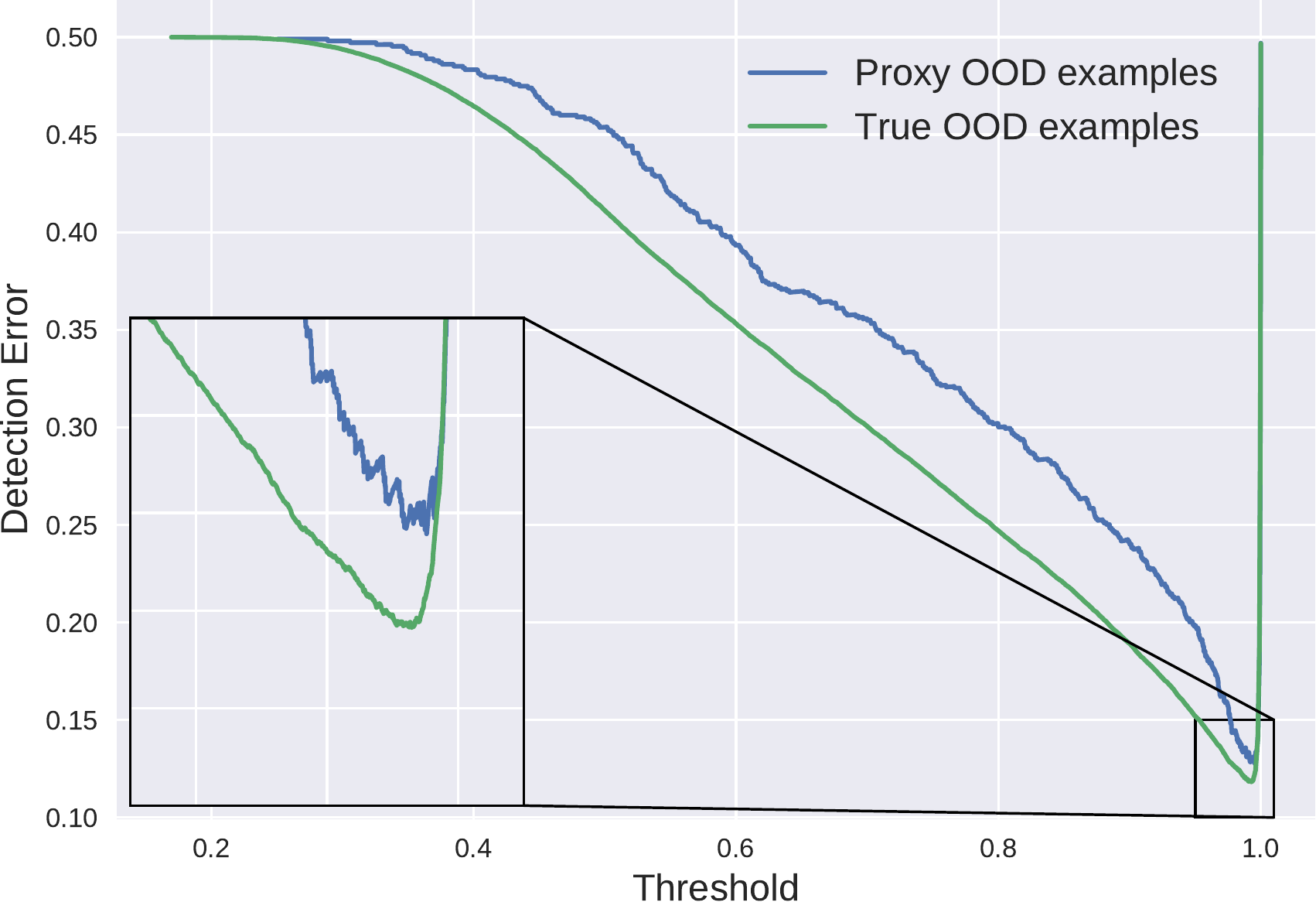}
 \subcaption{VGG13 (baseline)}
\end{minipage}\\

\begin{minipage}{0.30\textwidth}
 \includegraphics[width=\linewidth, trim={0cm 0cm 0cm 0cm}, clip]{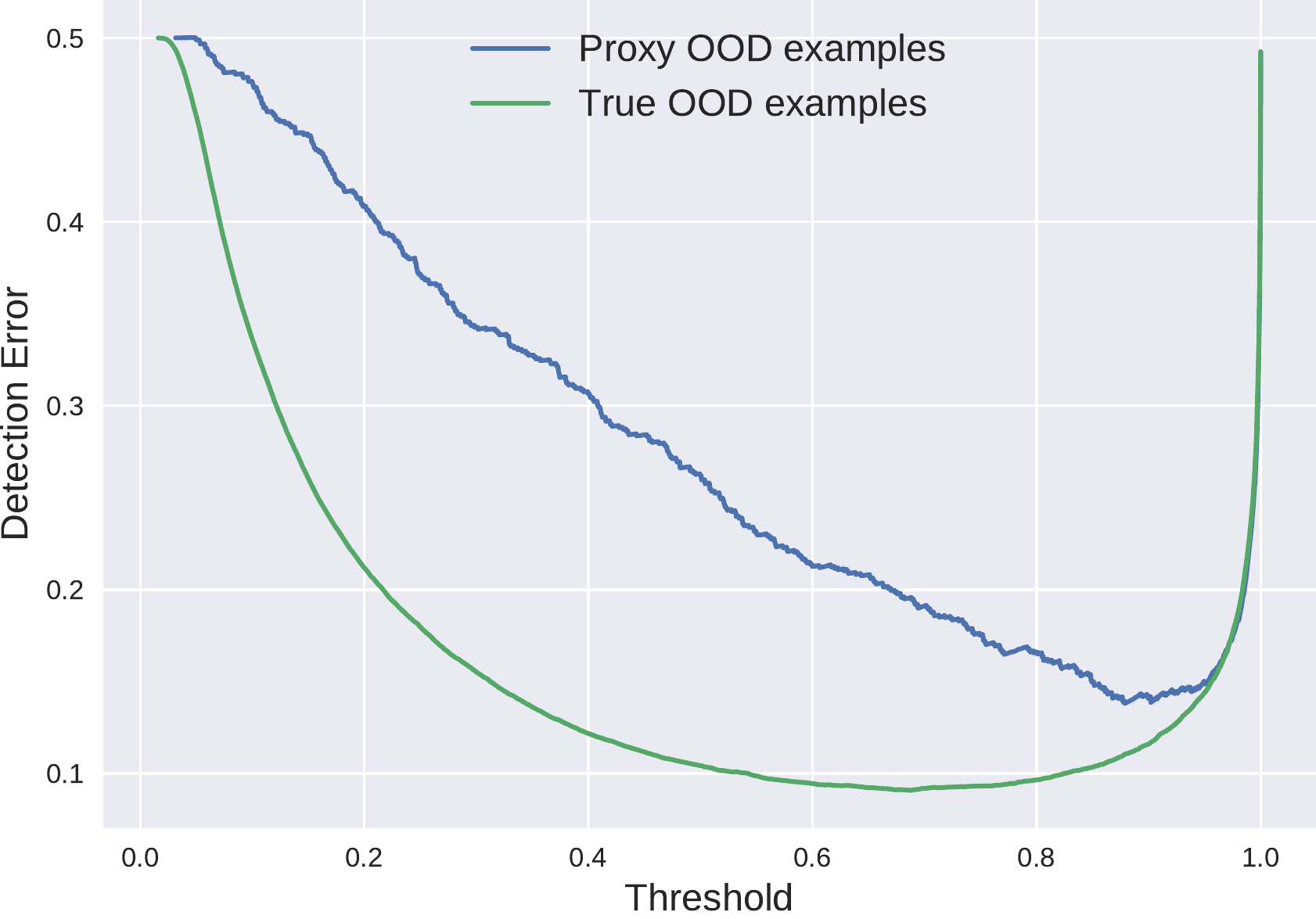}
 \subcaption{DenseNet (confidence)}
\end{minipage}
\begin{minipage}{0.30\textwidth}
 \includegraphics[width=\linewidth, trim={0cm 0cm 0cm 0cm}, clip]{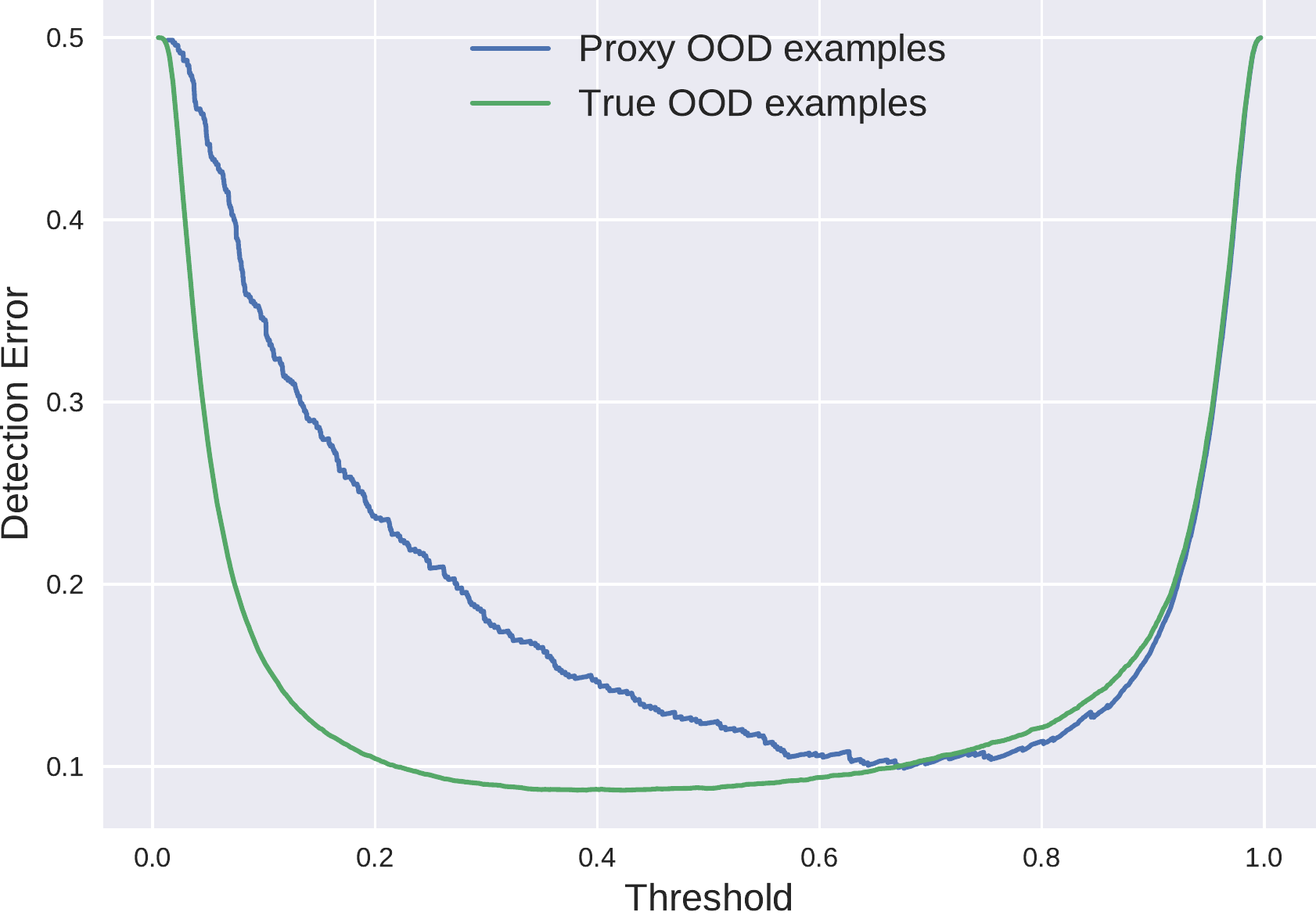}
 \subcaption{WideResNet (confidence)}
\end{minipage}
\begin{minipage}{0.30\textwidth}
 \includegraphics[width=\linewidth, trim={0cm 0cm 0cm 0cm}, clip]{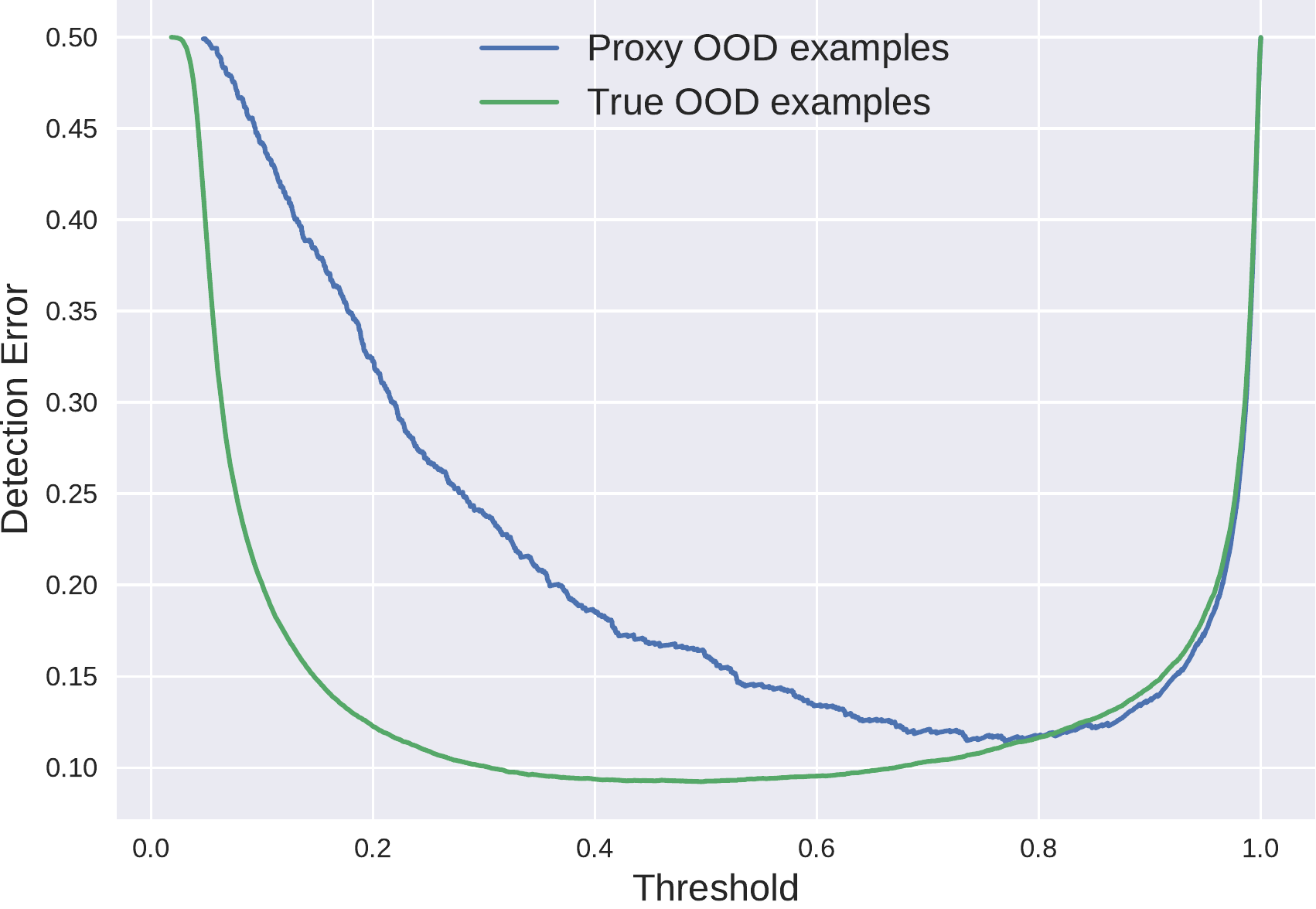}
 \subcaption{VGG13 (confidence)}
\end{minipage}\\
\caption{Detection error evaluated with respect to varying thresholds ($\delta$). Green lines depict detection error when the CIFAR-10 test set is in-distribution and the ``All Images'' dataset (TinyImageNet, LSUN, and iSUN) is out-of-distribution (OOD). Blue lines depict detection error when the correctly classified examples from the CIFAR-10 test set are in-distribution and the misclassified examples are out-of-distribution.}
\label{fig:detection_error_plots}
\end{figure*}

To visualize this case, we plot the detection error for in-distribution versus out-of-distribution examples for a range of detection thresholds (Figure~\ref{fig:detection_error_plots}). We find that misclassified examples yield similar detection error curves compared to true out-of-distribution examples, and as a result, the optimal detection threshold for separating correct and incorrectly classified in-distribution examples may also be used as the threshold for out-of-distribution detection. We evaluate this empirically in Table~\ref{table:calibrate_on_misclassified}, which demonstrates that in most cases, the performance of the out-of-distribution detector is reduced, but is still acceptable, when using the misclassification threshold. We note that it may be practical to use the misclassification threshold in any case, as the misclassification threshold is observed to be greater than the optimal out-of-distribution detection threshold for all tests. As a result, we obtain a more conservative out-of-distribution detector, and an optimal misclassification detector.

\begin{table}[h]
\centering
\caption{Detection error when CIFAR-10 is in-distribution and ``All Images'' is out-of-distribution. The first column shows values when the detection error is calibrated on a holdout set of out-of-distribution examples (reproduced from Table~\ref{table:baseline_confidence} for convenience). The second column shows values when detection error is calibrated using misclassified in-distribution examples as a proxy for out-of-distribution examples. Results averaged over 5 runs.}
\vspace{0.1in}
\begin{tabular}{lcc}
\hline
\multicolumn{1}{c}{\textbf{Architecture}} & \textbf{\begin{tabular}[c]{@{}c@{}}Detection Error\\ (calibrated on\\ true OOD)\end{tabular}} & \textbf{\begin{tabular}[c]{@{}c@{}}Detection Error\\ (calibrated on\\ misclassified)\end{tabular}} \\
\hline
& \multicolumn{2}{c}{\textbf{Baseline/Confidence}} \\
 \cline{2-3}
DenseNet-BC & 11.6/\textbf{10.9} & \textbf{11.8}/12.7 \\
WRN-28-10 & 12.4/\textbf{9.7} & 12.5/\textbf{10.6} \\
VGG13 & 11.7/\textbf{9.1} & 12.3/\textbf{10.8} \\
\hline
\end{tabular}
\label{table:calibrate_on_misclassified}
\end{table}

\section{Related Work}
Our work on confidence estimation is similar in some ways to concurrently developed work on uncertainty estimation for regression tasks by \citet{KendallGal2017UncertaintiesB} and \citet{gurevich2017learning}. Similar to our model architecture, they both train neural networks that produce two outputs: a prediction and an uncertainty estimate. The formulation of their loss functions is also similar to ours, as both down-weight the loss on the main prediction task with respect to the amount of uncertainty, at the cost of some penalty. Although similar to the implementation of our confidence estimation technique, these formulations do not transfer well to classification tasks.

To address this issue, \citet{KendallGal2017UncertaintiesB} also present a separate uncertainty estimation technique for classification tasks. Here, they encourage their model to learn uncertainty by adding noise to the class prediction logits, where the magnitude of the noise is proportional to the estimated uncertainty for the given input. Their measure differs to ours, in that the uncertainty estimates yield unbounded values, while our confidence estimates produce normalized values. While both techniques enable a measure of potential risk, the latter may be preferable for human-in-the-loop applications.

\section{Conclusion}
In this paper, we introduce an intuitive technique for learning confidence estimates for neural networks, without requiring any labels for supervised training. We evaluate the utility of the predicted confidence on the task of out-of-distribution detection, where we demonstrate  improvements over output thresholding \citep{hendrycks2016baseline} and ODIN \citep{anonymous2018enhancing}. We also demonstrate that misclassified examples can be used to calibrate out-of-distribution detectors. In the future, we intend to apply learned confidence estimates to tasks beyond classification, such as semantic segmentation and natural language understanding. We also intend to experiment with more complex and human-inspired schemes for giving ``hints''.

\clearpage

\bibliography{example_paper}
\bibliographystyle{icml2018}

\appendix
\label{appendix}
\onecolumn
\section{Supplementary Materials}

\begin{figure*}[!htbp]
\centering
\begin{minipage}{0.22\textwidth}
 \includegraphics[width=\linewidth, trim={0.75cm 0.75cm 1.5cm 0cm}, clip]{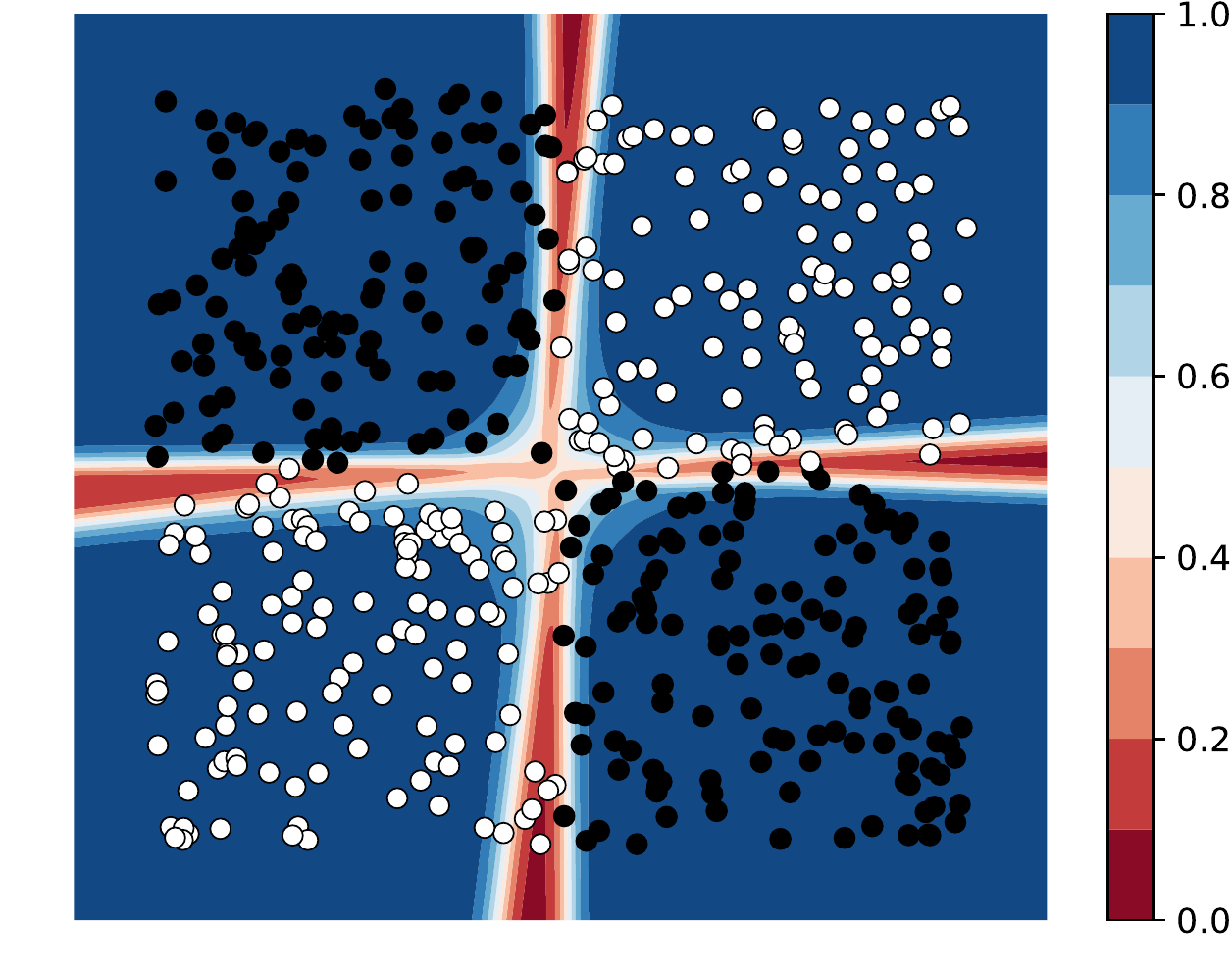}
 \subcaption{$\beta$ = 0.2, noise = 0\%}
\end{minipage}
\begin{minipage}{0.22\textwidth}
 \includegraphics[width=\linewidth, trim={0.75cm 0.75cm 1.5cm 0cm}, clip]{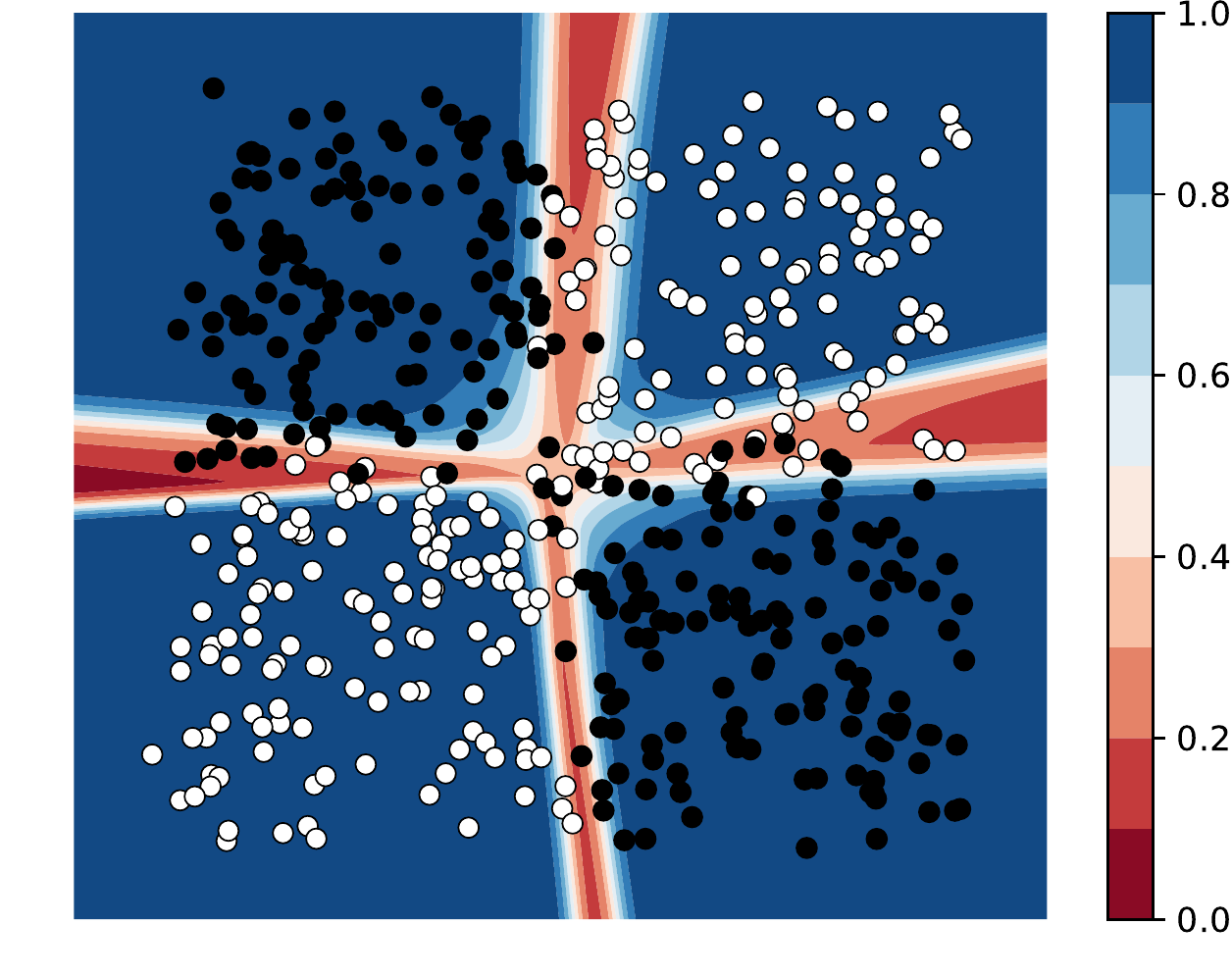}
 \subcaption{$\beta$ = 0.2, noise = 10\%}
\end{minipage}
\begin{minipage}{0.22\textwidth}
 \includegraphics[width=\linewidth, trim={0.75cm 0.75cm 1.5cm 0cm}, clip]{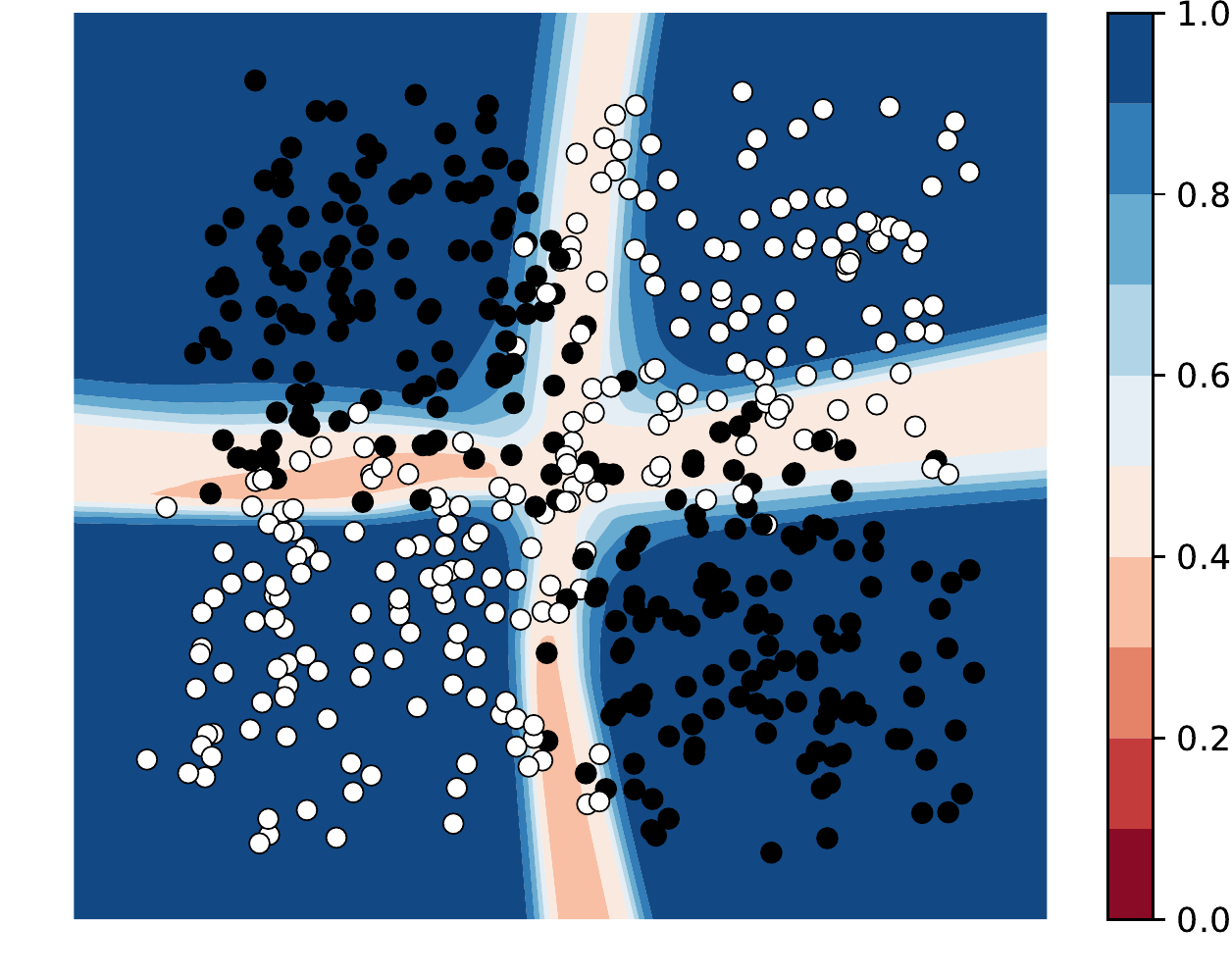}
 \subcaption{$\beta$ = 0.2, noise = 20\%}
\end{minipage}
\begin{minipage}{0.247\textwidth}
 \includegraphics[width=\linewidth, trim={0.75cm 0.5cm 0cm 0cm}, clip]{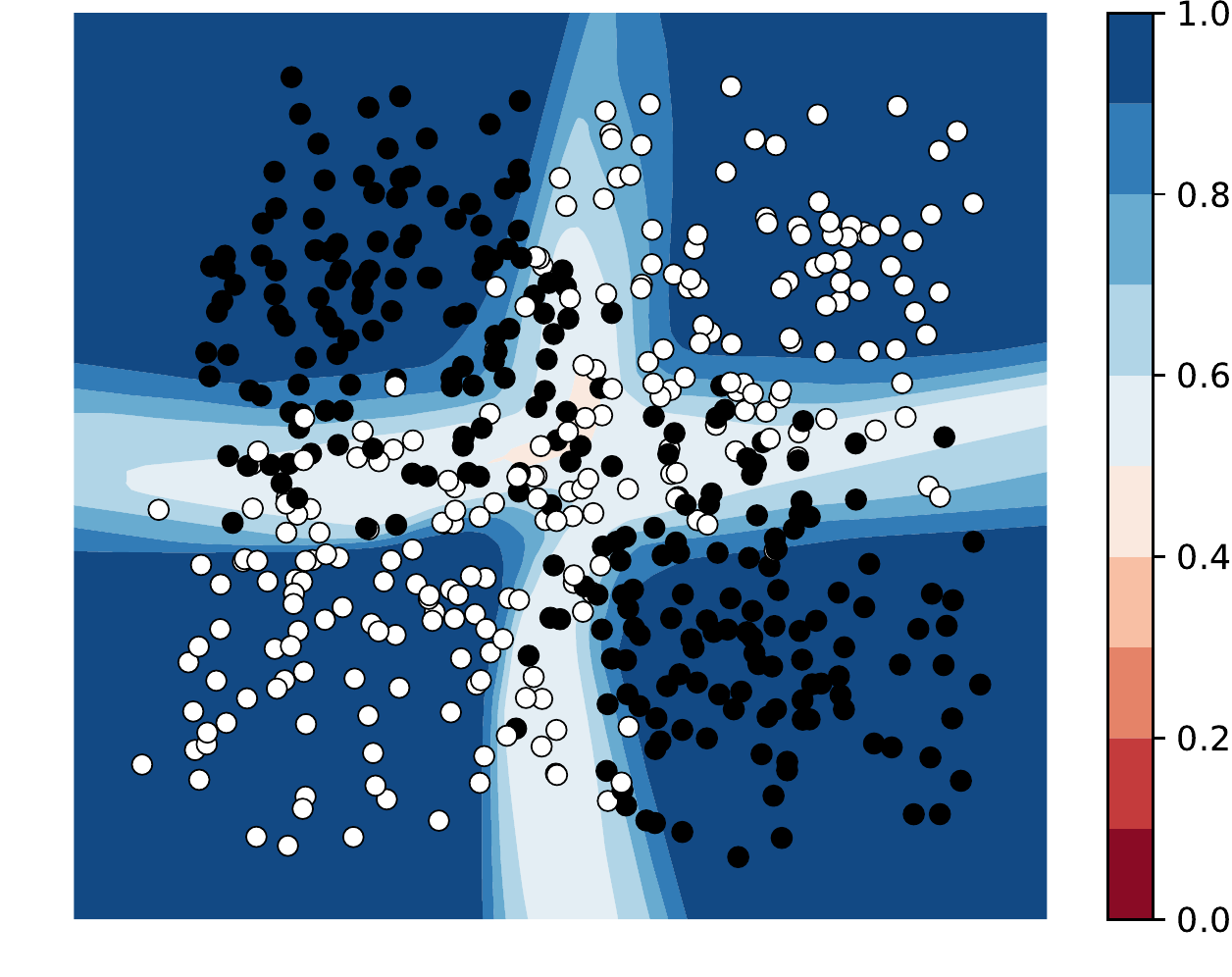}
 \subcaption{$\beta$ = 0.2, noise = 30\%}
\end{minipage}\\

\begin{minipage}{0.22\textwidth}
 \includegraphics[width=\linewidth, trim={0.75cm 0.75cm 1.5cm 0cm}, clip]{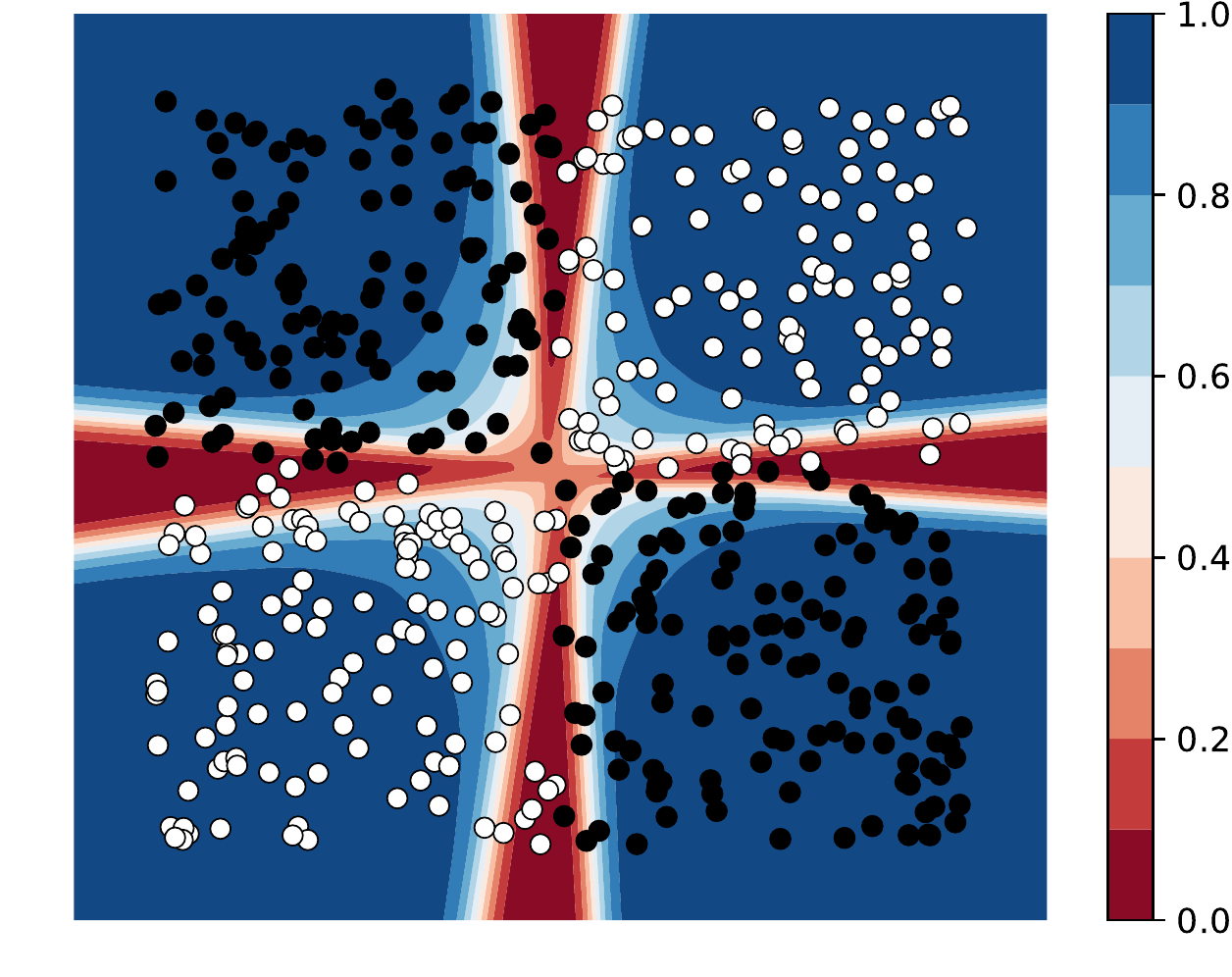}
 \subcaption{$\beta$ = 0.4, noise = 0\%}
\end{minipage}
\begin{minipage}{0.22\textwidth}
 \includegraphics[width=\linewidth, trim={0.75cm 0.75cm 1.5cm 0cm}, clip]{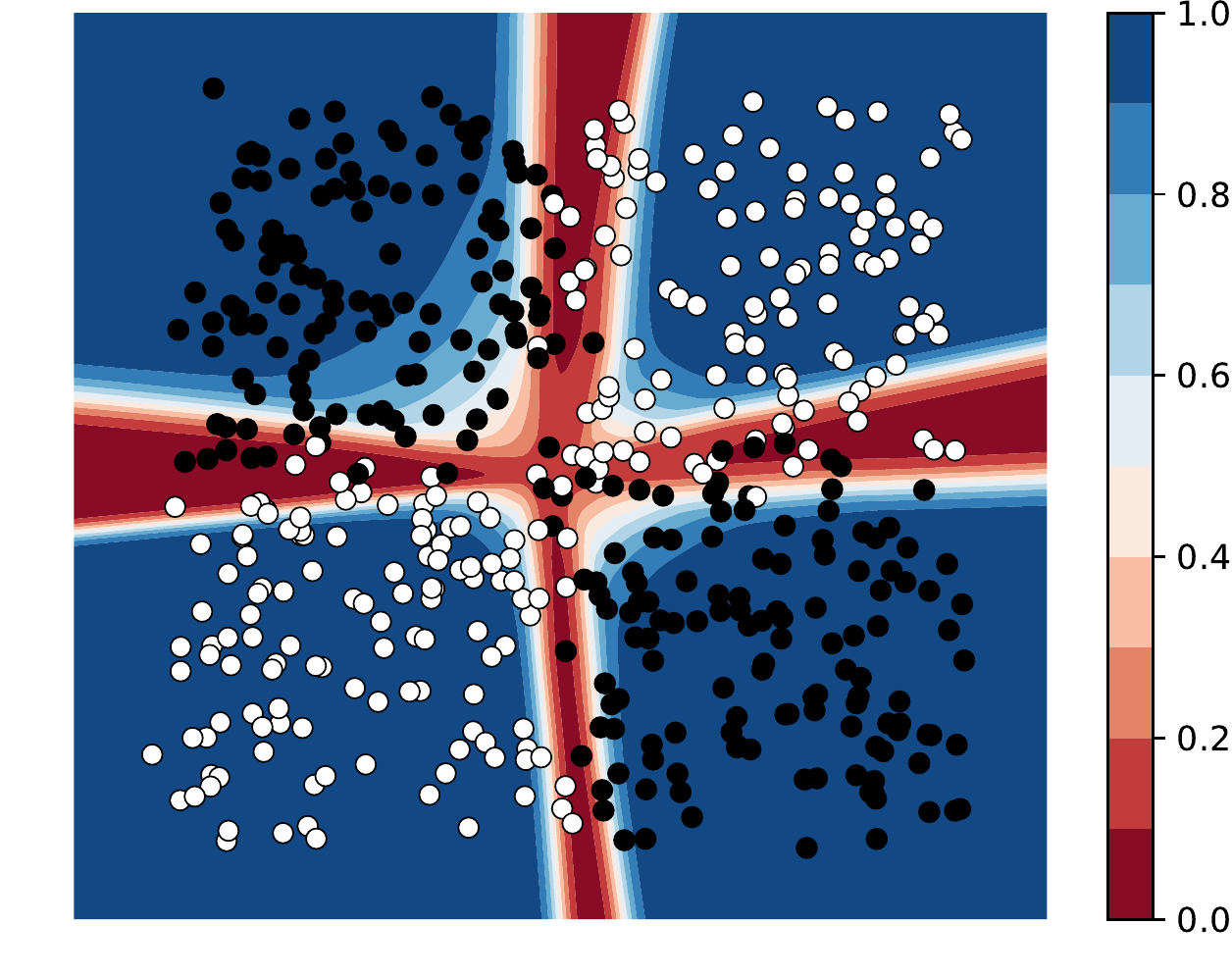}
 \subcaption{$\beta$ = 0.4, noise = 10\%}
\end{minipage}
\begin{minipage}{0.22\textwidth}
 \includegraphics[width=\linewidth, trim={0.75cm 0.75cm 1.5cm 0cm}, clip]{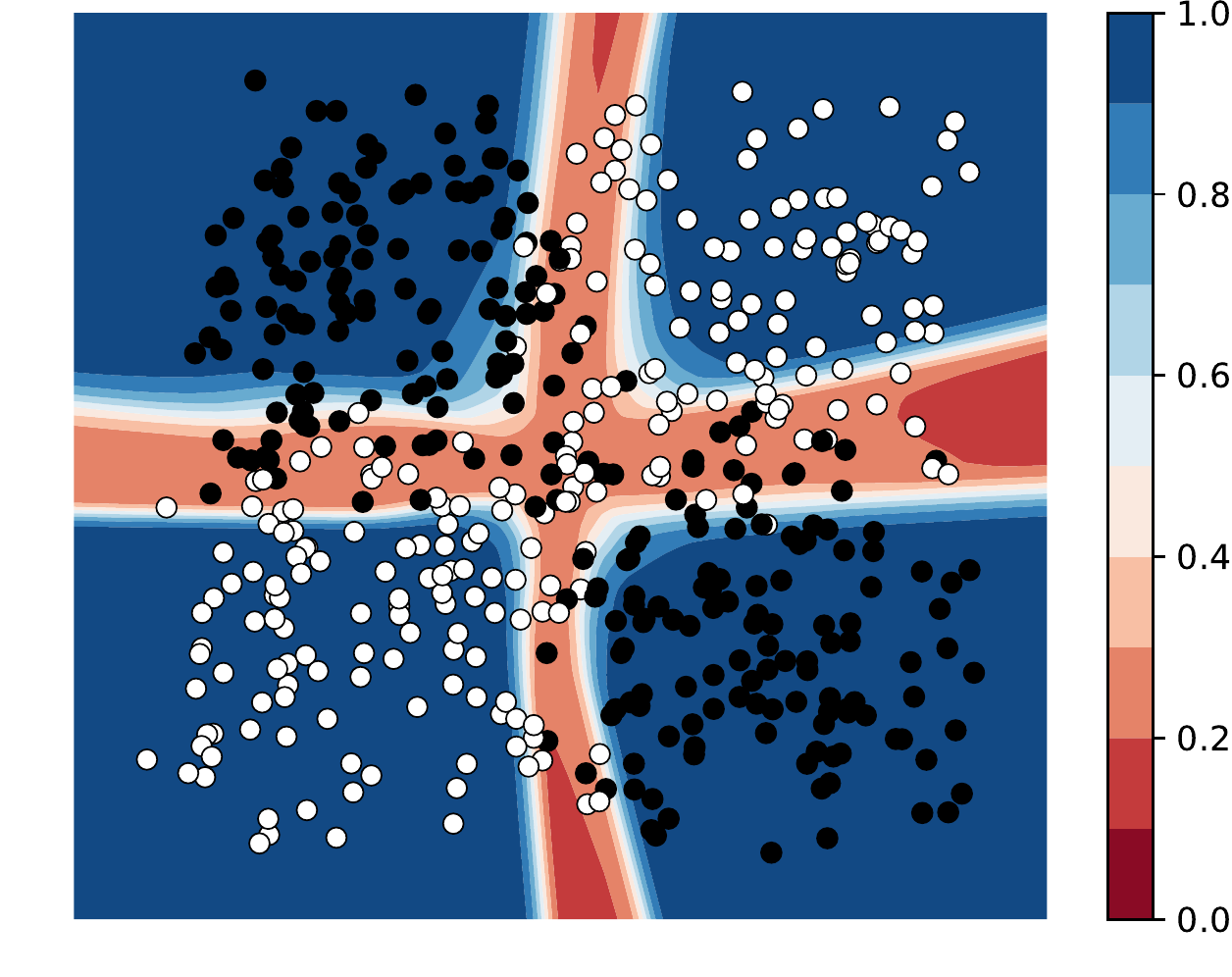}
 \subcaption{$\beta$ = 0.4, noise = 20\%}
\end{minipage}
\begin{minipage}{0.247\textwidth}
 \includegraphics[width=\linewidth, trim={0.75cm 0.5cm 0cm 0cm}, clip]{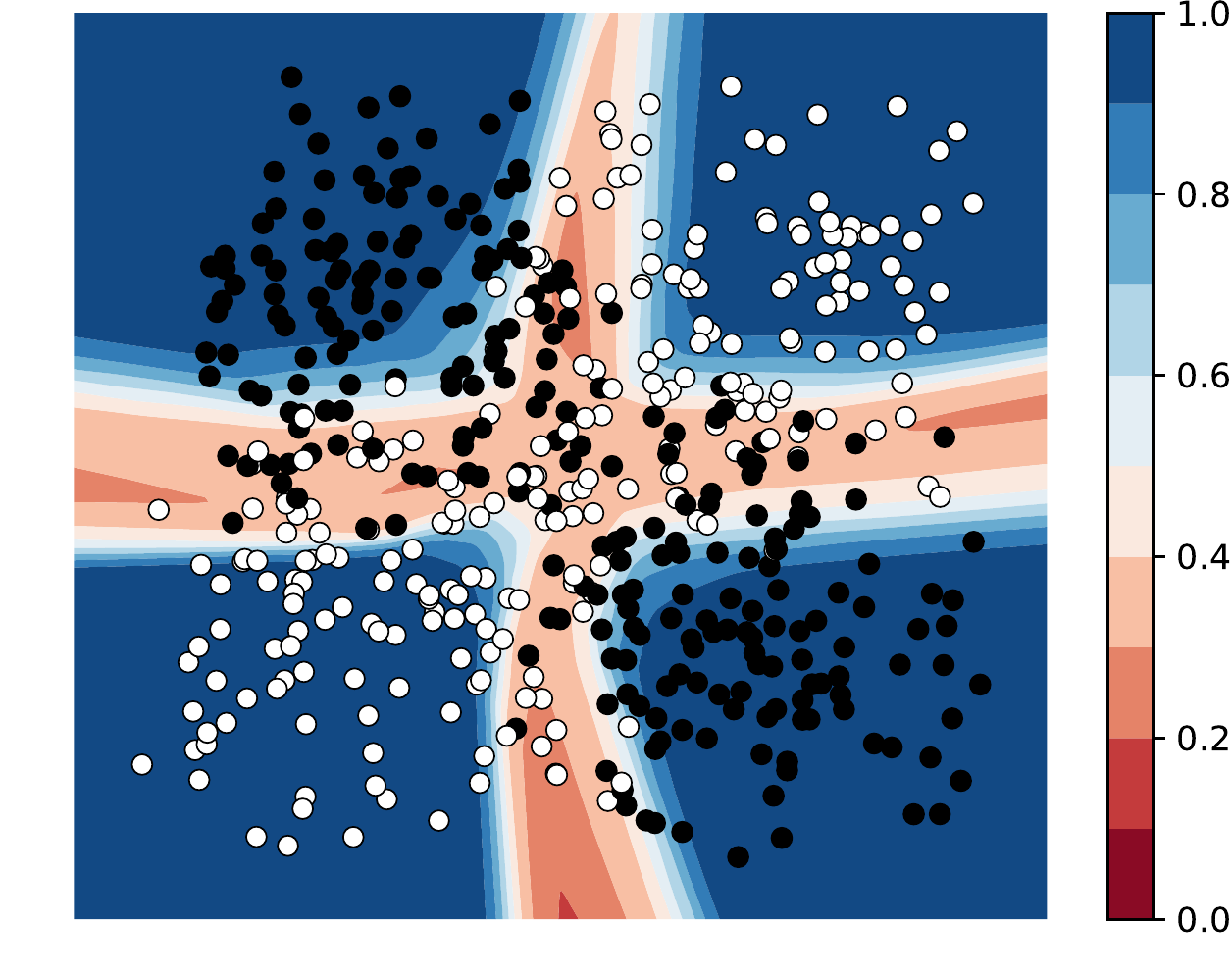}
 \subcaption{$\beta$ = 0.4, noise = 30\%}
\end{minipage}\\

\begin{minipage}{0.22\textwidth}
 \includegraphics[width=\linewidth, trim={0.75cm 0.75cm 1.5cm 0cm}, clip]{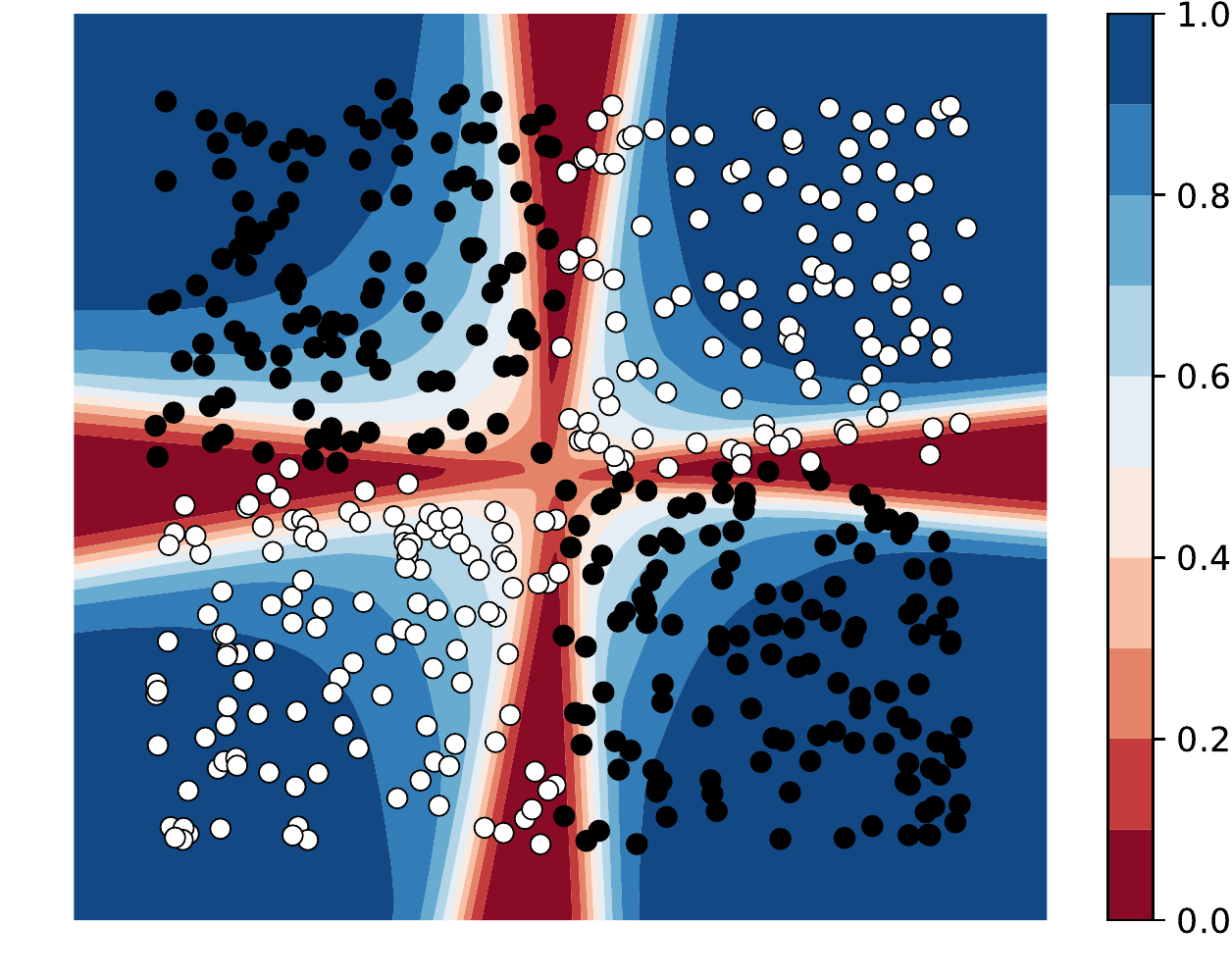}
 \subcaption{$\beta$ = 0.6, noise = 0\%}
\end{minipage}
\begin{minipage}{0.22\textwidth}
 \includegraphics[width=\linewidth, trim={0.75cm 0.75cm 1.5cm 0cm}, clip]{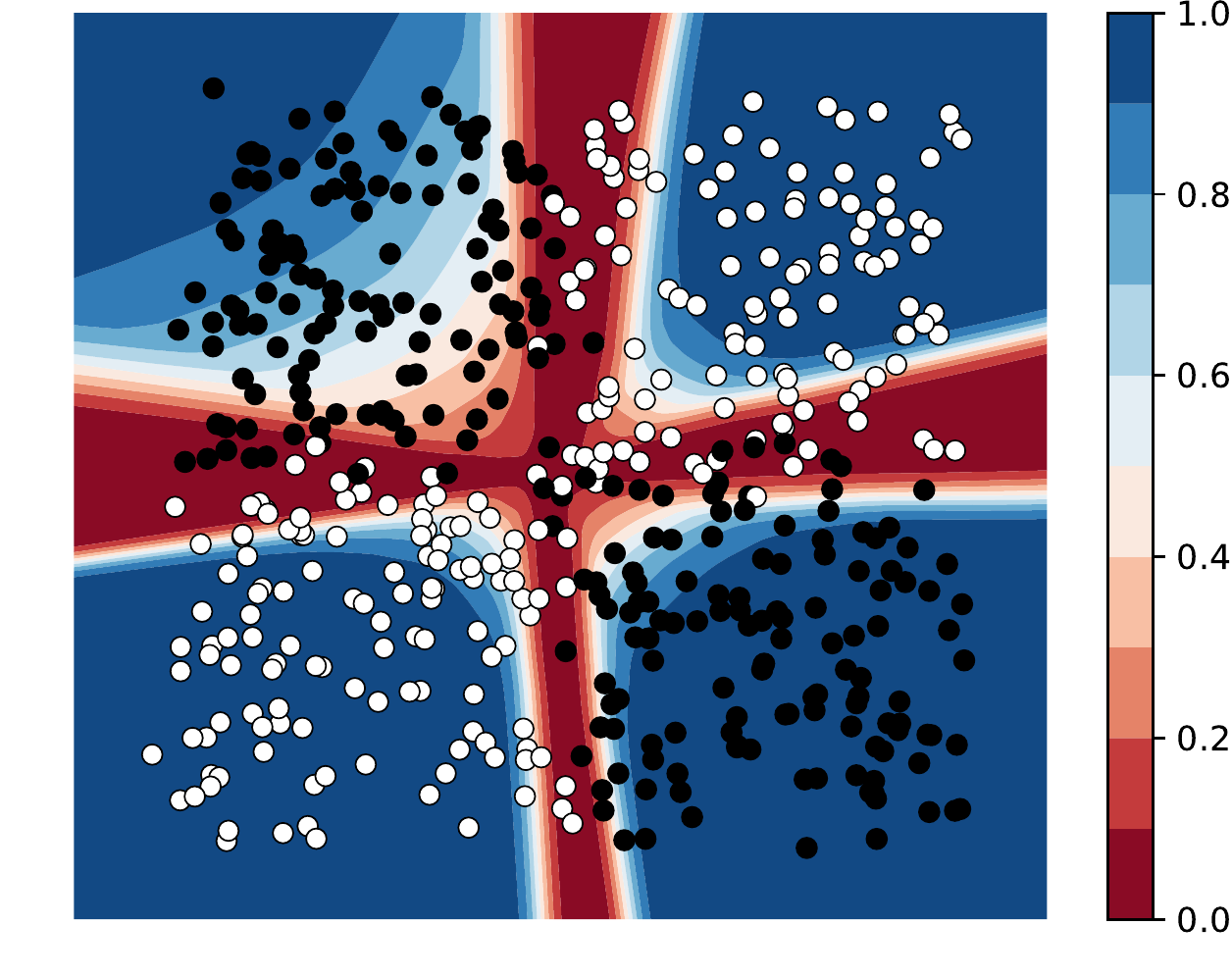}
 \subcaption{$\beta$ = 0.6, noise = 10\%}
\end{minipage}
\begin{minipage}{0.22\textwidth}
 \includegraphics[width=\linewidth, trim={0.75cm 0.75cm 1.5cm 0cm}, clip]{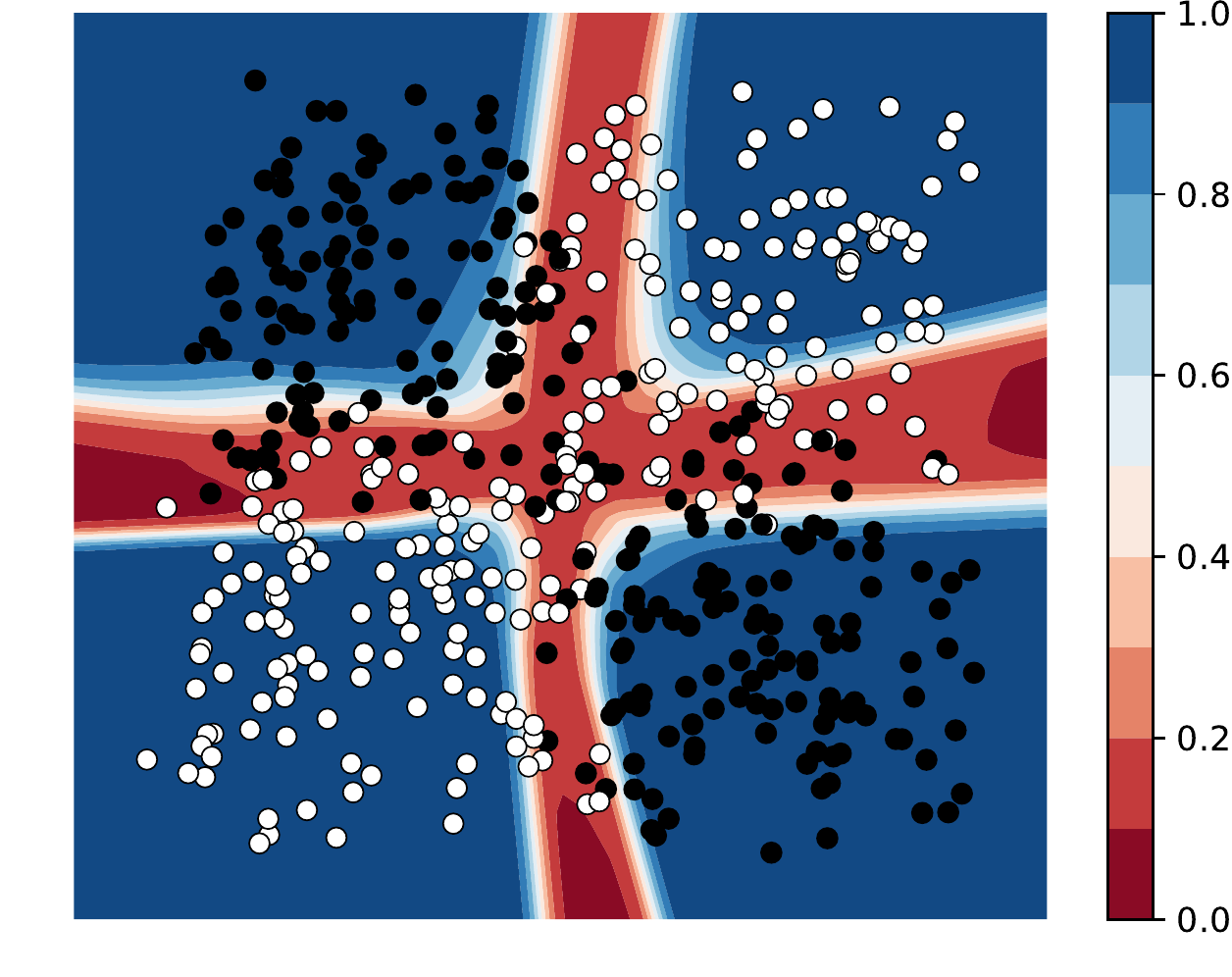}
 \subcaption{$\beta$ = 0.6, noise = 20\%}
\end{minipage}
\begin{minipage}{0.247\textwidth}
 \includegraphics[width=\linewidth, trim={0.75cm 0.5cm 0cm 0cm}, clip]{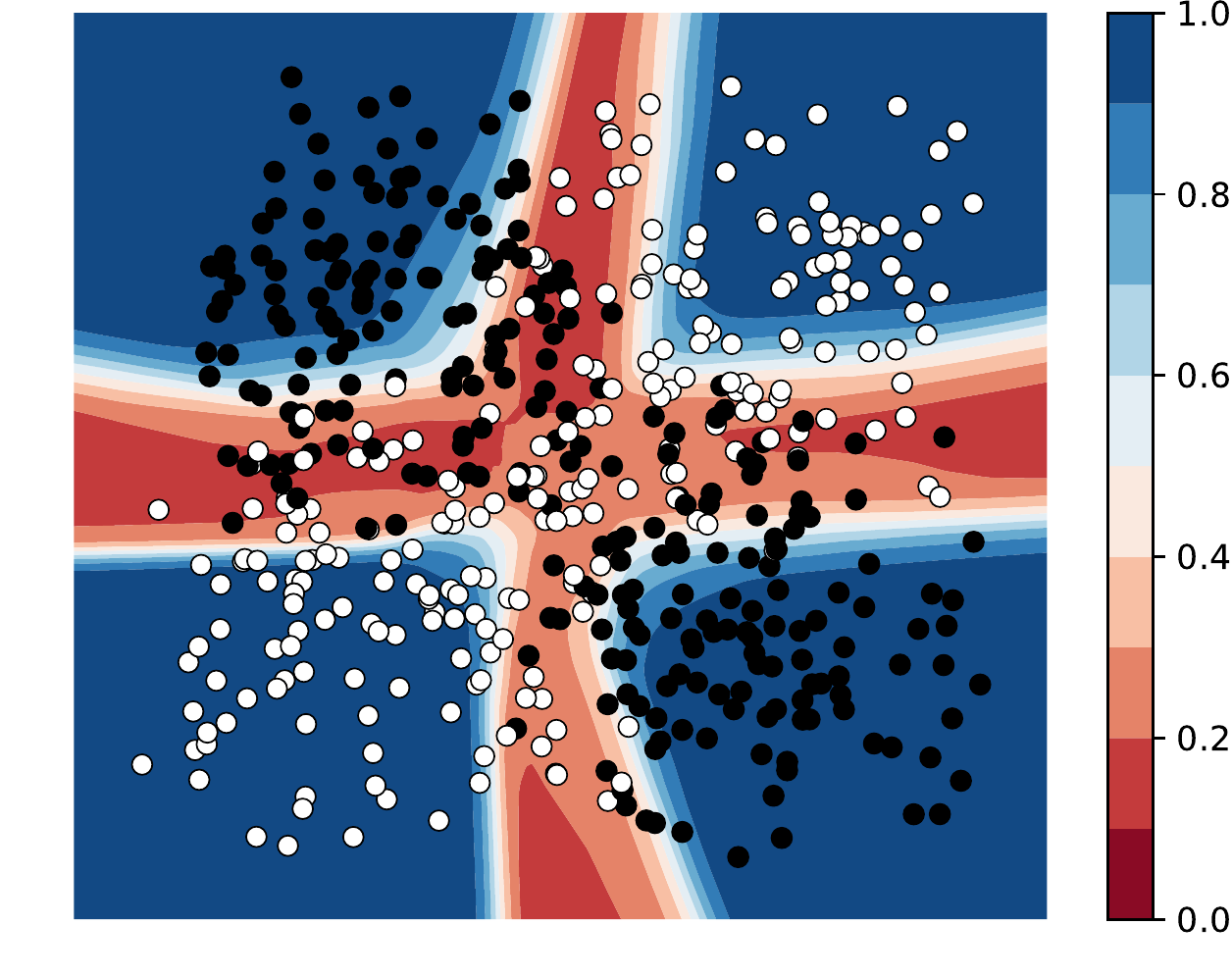}
 \subcaption{$\beta$ = 0.6, noise = 30\%}
\end{minipage}\\

\begin{minipage}{0.22\textwidth}
 \includegraphics[width=\linewidth, trim={0.75cm 0.75cm 1.5cm 0cm}, clip]{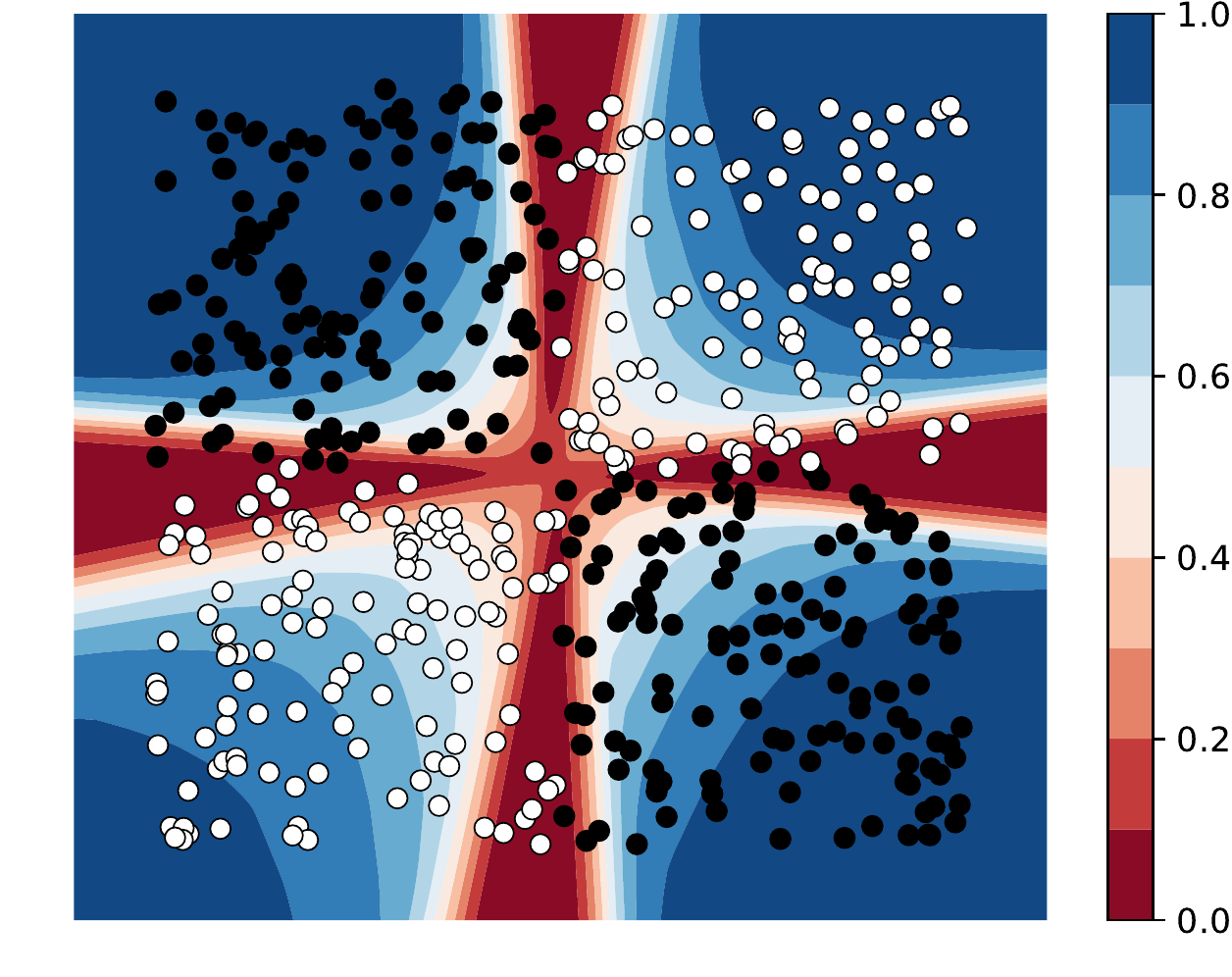}
 \subcaption{$\beta$ = 0.8, noise = 0\%}
\end{minipage}
\begin{minipage}{0.22\textwidth}
 \includegraphics[width=\linewidth, trim={0.75cm 0.75cm 1.5cm 0cm}, clip]{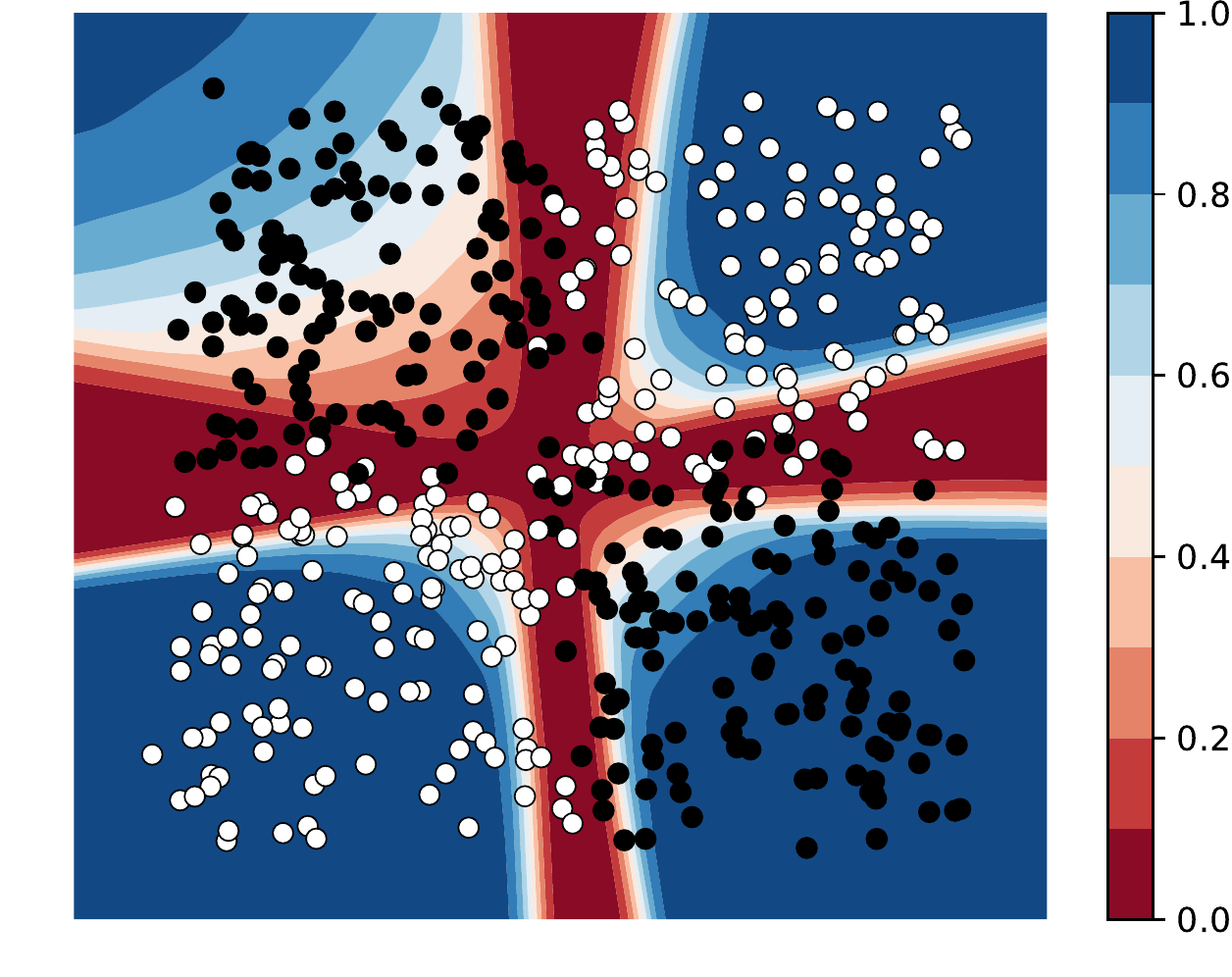}
 \subcaption{$\beta$ = 0.8, noise = 10\%}
\end{minipage}
\begin{minipage}{0.22\textwidth}
 \includegraphics[width=\linewidth, trim={0.75cm 0.75cm 1.5cm 0cm}, clip]{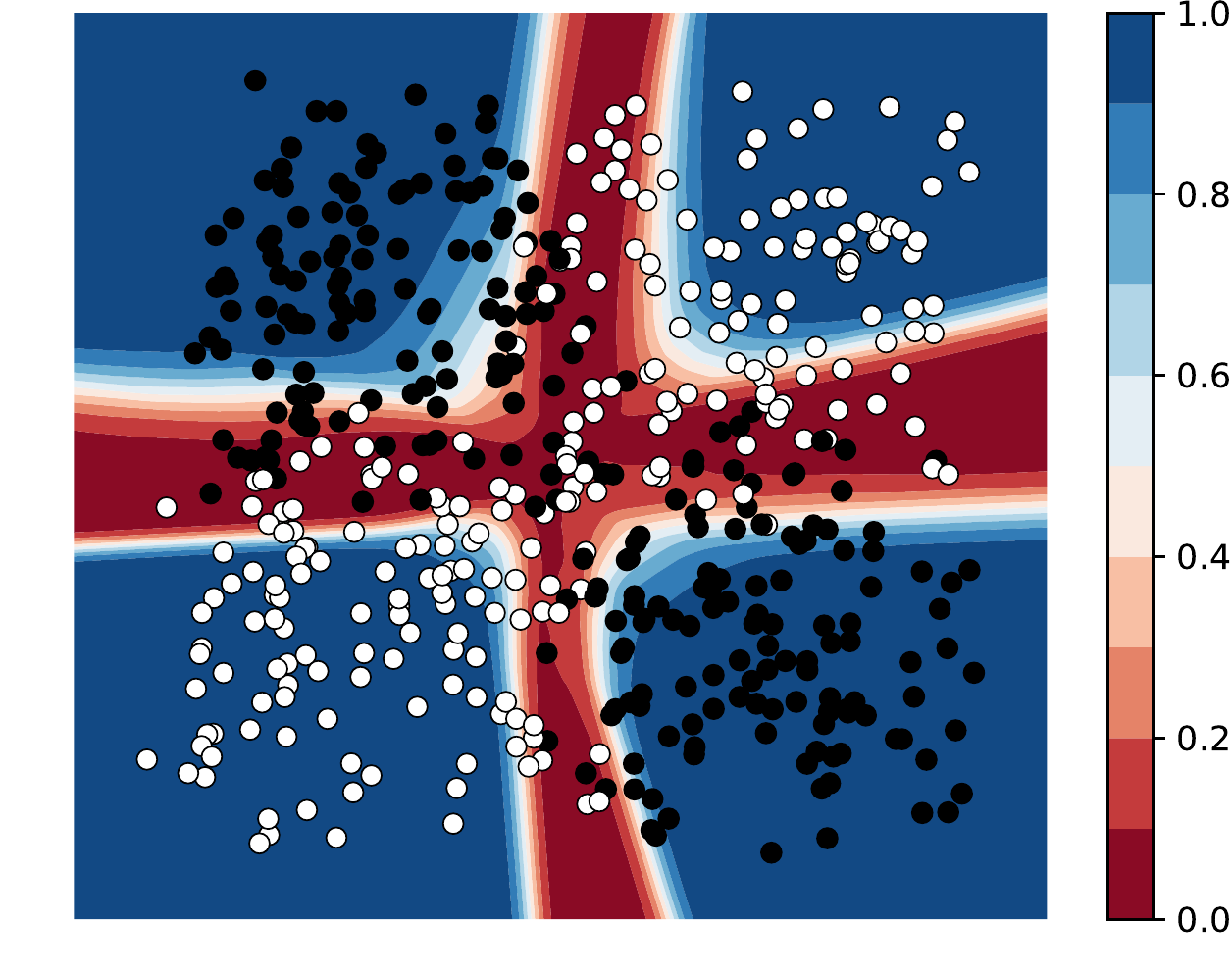}
 \subcaption{$\beta$ = 0.8, noise = 20\%}
\end{minipage}
\begin{minipage}{0.247\textwidth}
 \includegraphics[width=\linewidth, trim={0.75cm 0.5cm 0cm 0cm}, clip]{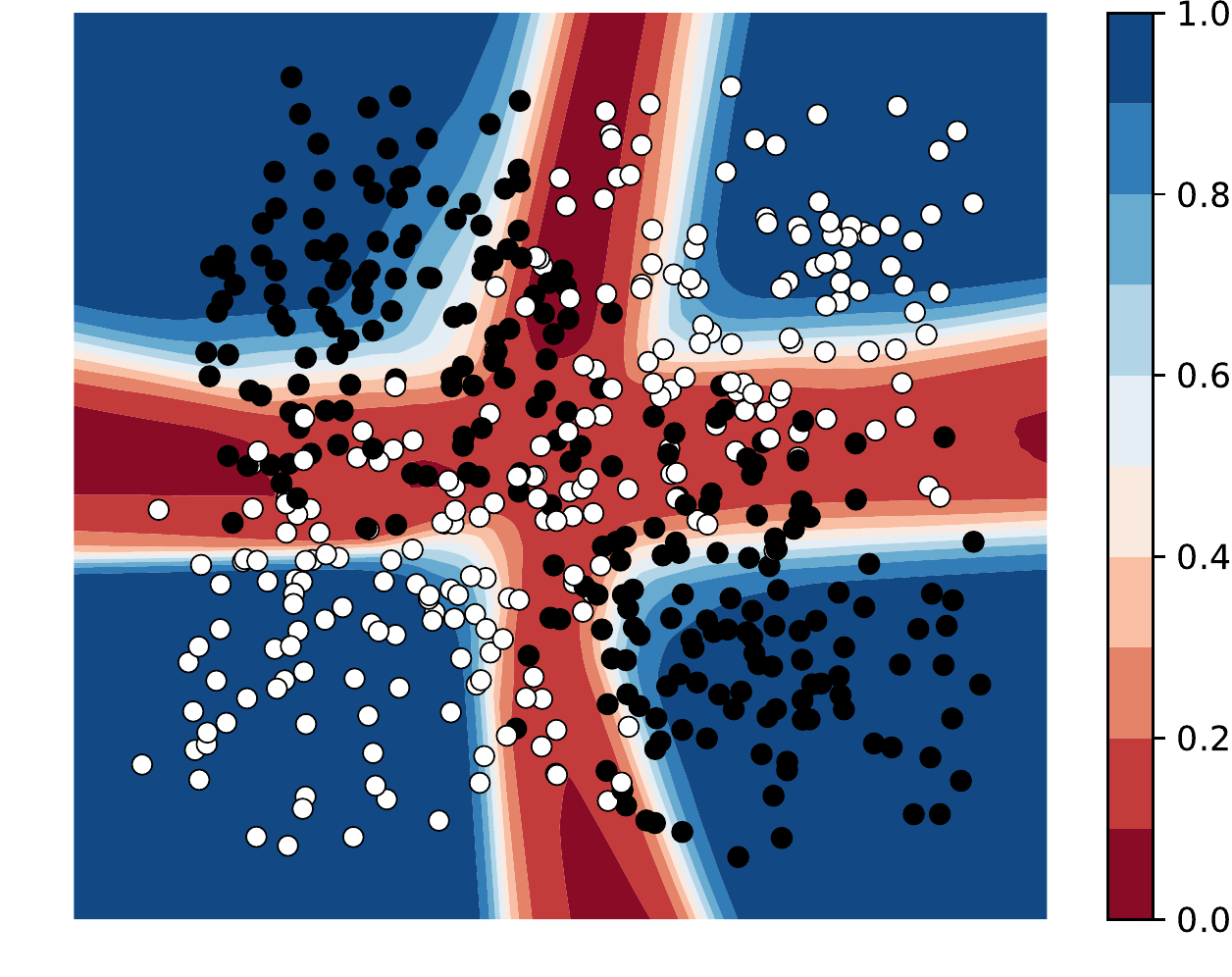}
 \subcaption{$\beta$ = 0.8, noise = 30\%}
\end{minipage}\\

\begin{minipage}{0.22\textwidth}
 \includegraphics[width=\linewidth, trim={0.75cm 0.75cm 1.5cm 0cm}, clip]{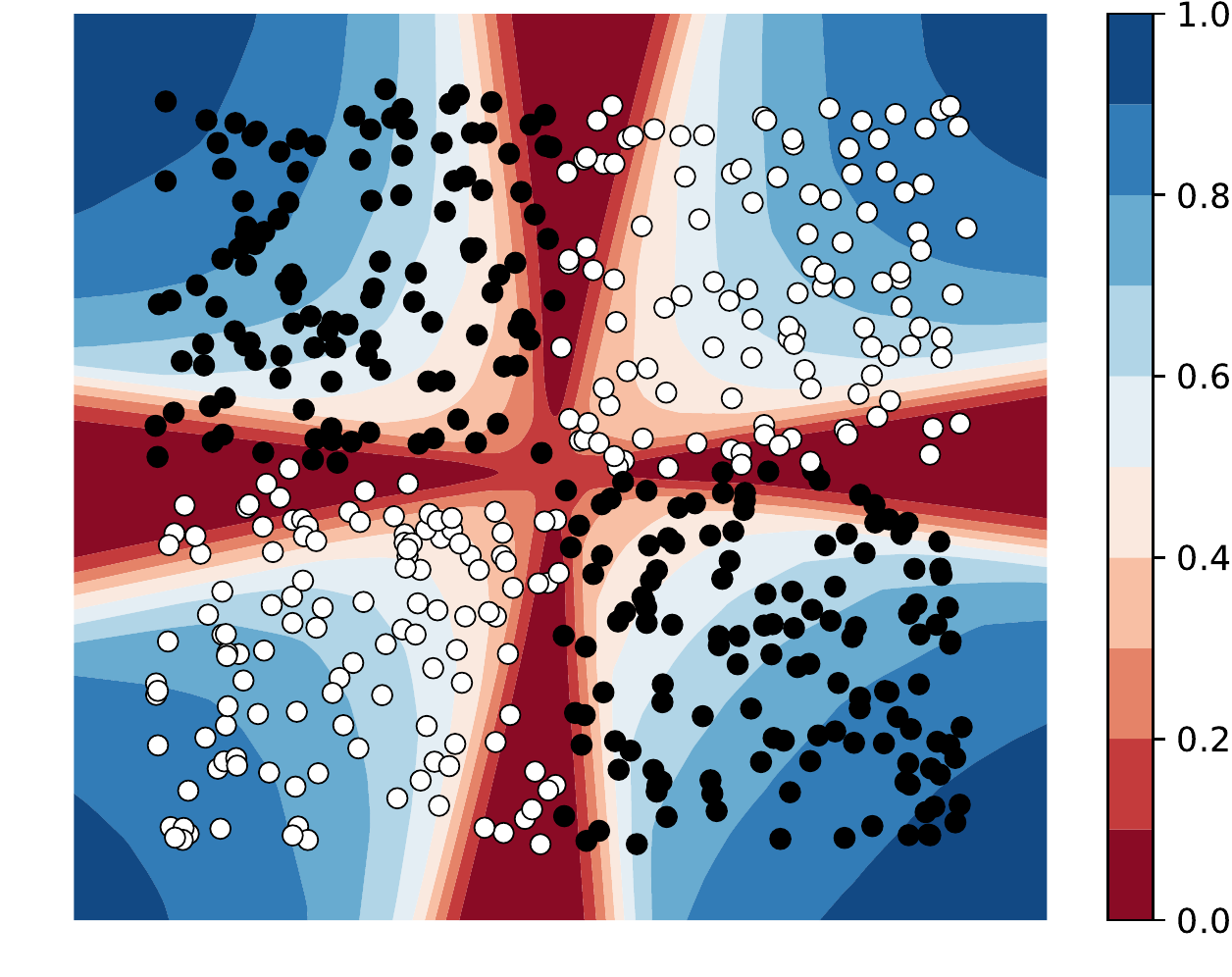}
 \subcaption{$\beta$ = 1.0, noise = 0\%}
\end{minipage}
\begin{minipage}{0.22\textwidth}
 \includegraphics[width=\linewidth, trim={0.75cm 0.75cm 1.5cm 0cm}, clip]{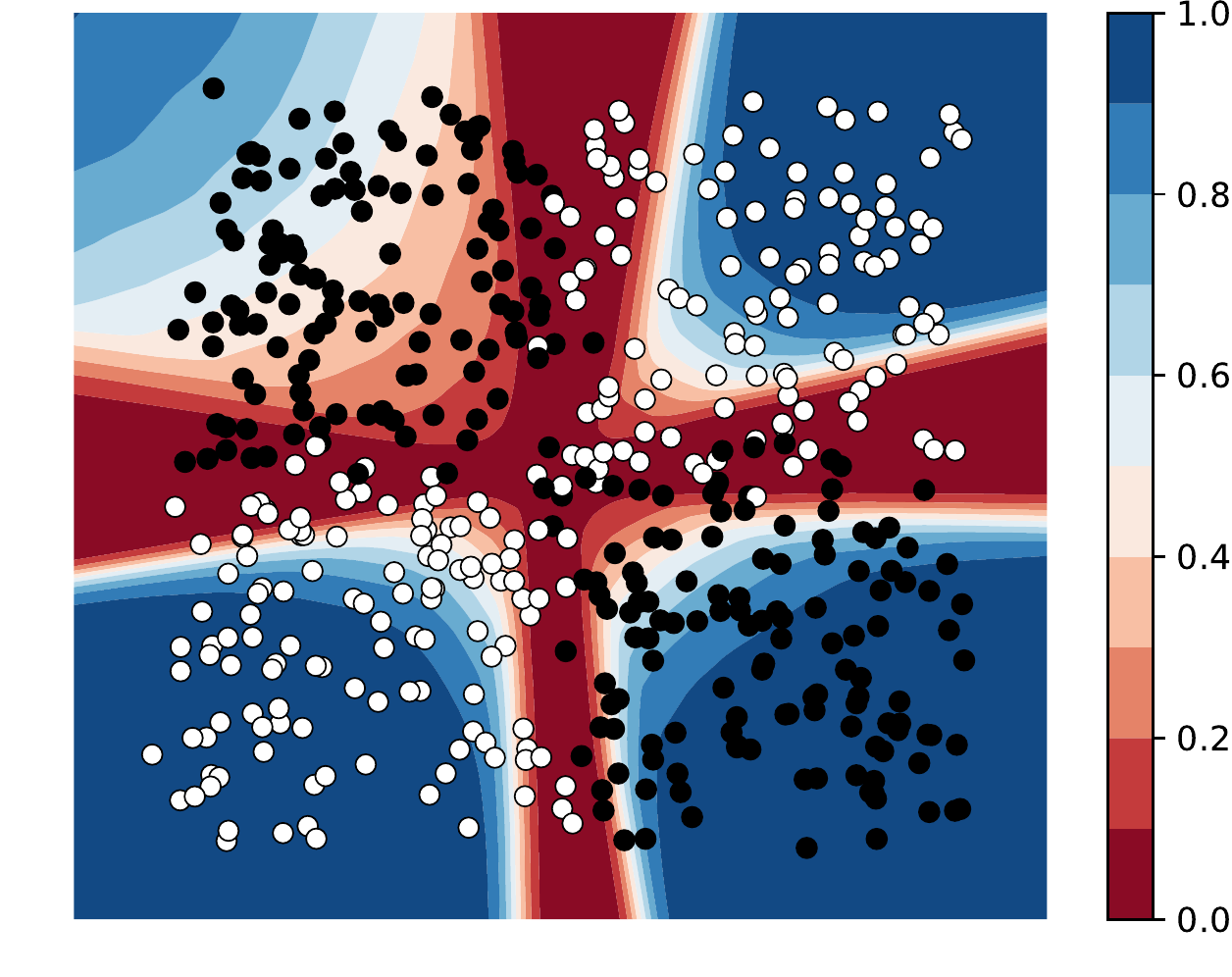}
 \subcaption{$\beta$ = 1.0, noise = 10\%}
\end{minipage}
\begin{minipage}{0.22\textwidth}
 \includegraphics[width=\linewidth, trim={0.75cm 0.75cm 1.5cm 0cm}, clip]{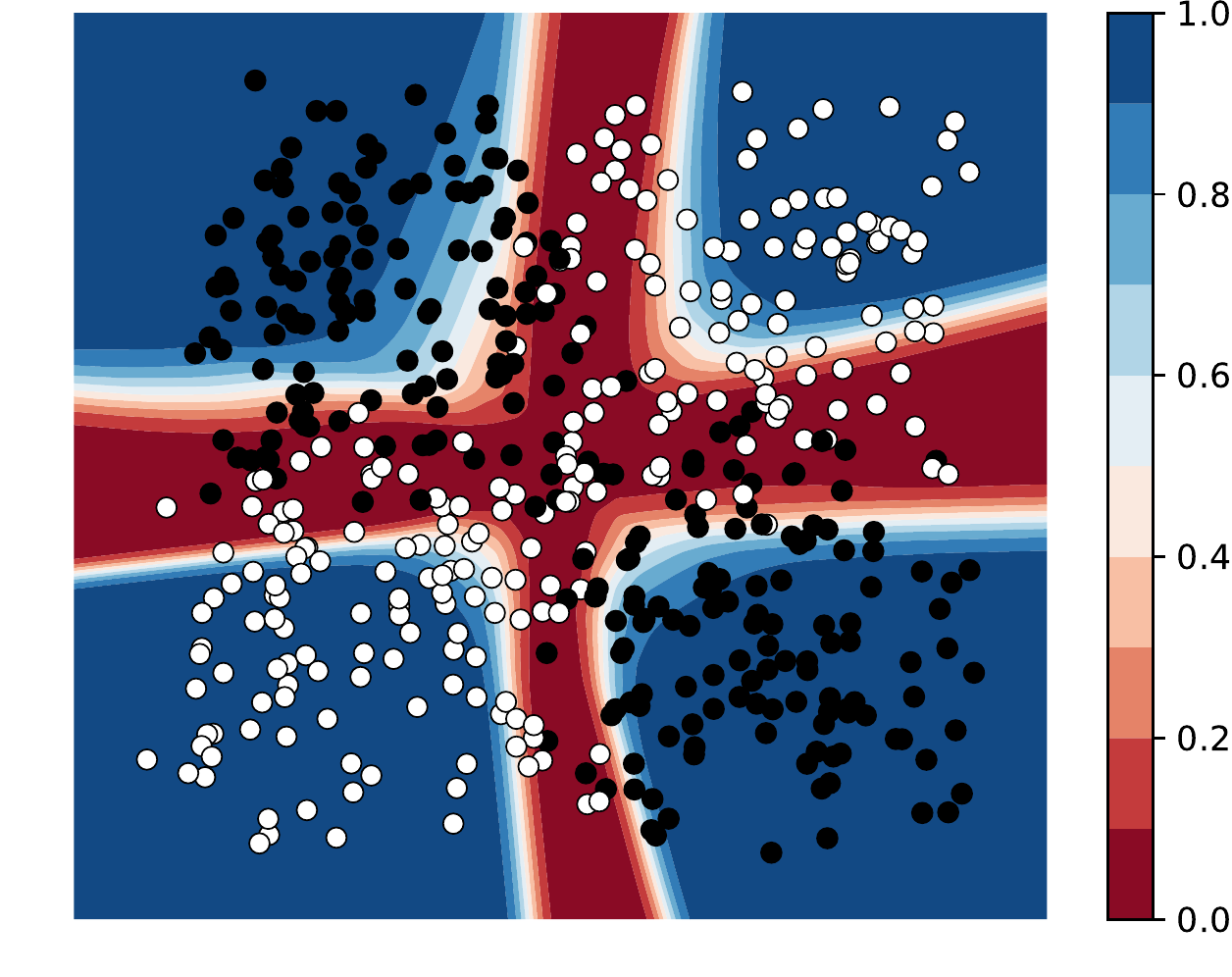}
 \subcaption{$\beta$ = 1.0, noise = 20\%}
\end{minipage}
\begin{minipage}{0.247\textwidth}
 \includegraphics[width=\linewidth, trim={0.75cm 0.5cm 0cm 0cm}, clip]{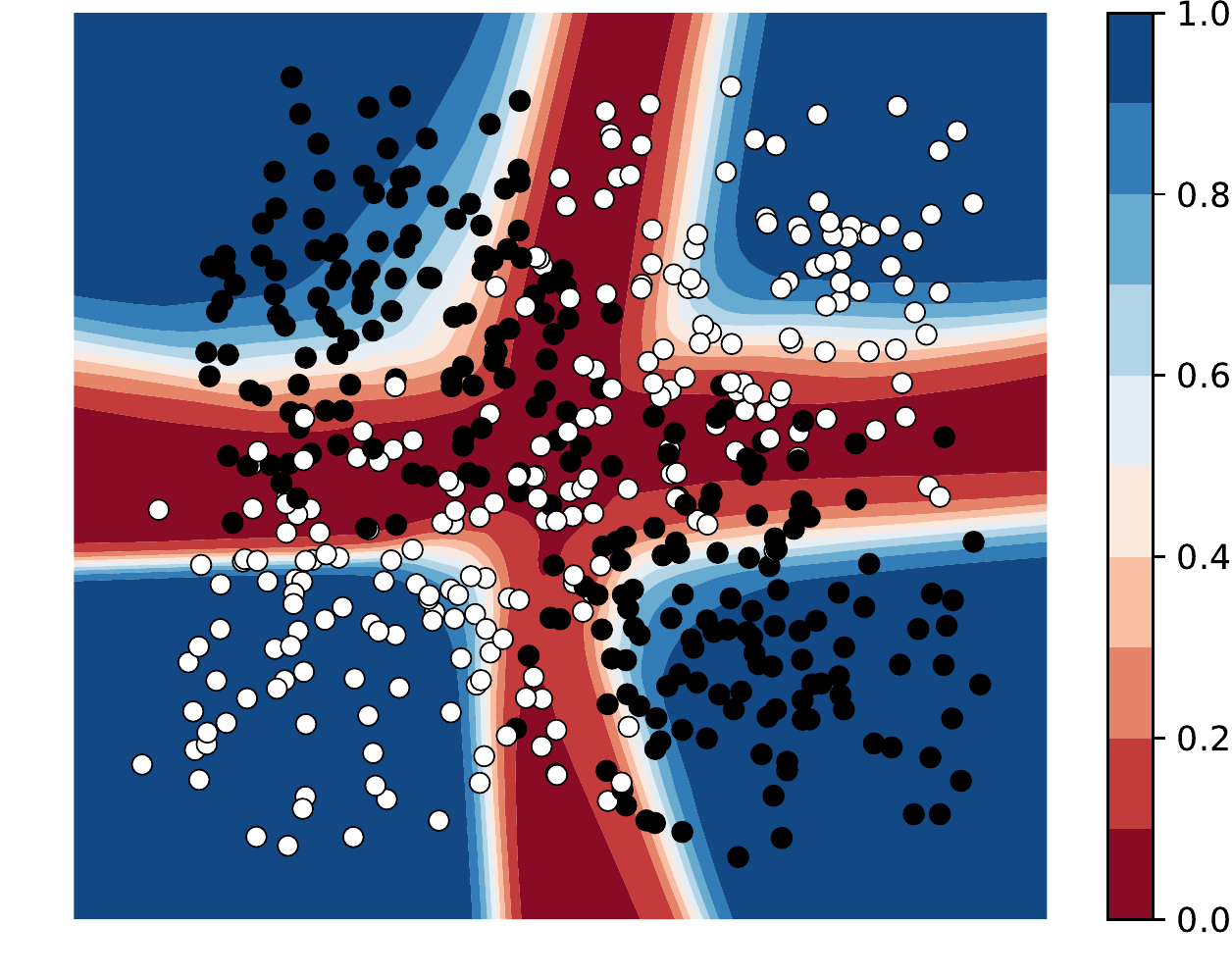}
 \subcaption{$\beta$ = 1.0, noise = 30\%}
\end{minipage}\\
\caption{Confidence predictions for XOR dataset. The amount of noise in the data increases from left to right, while the confidence budget $\beta$ increases from top to bottom. The training set is shown here to demonstrate how the confidence decision boundary tries to extend to cover most or all of the noisy regions.} 
\label{fig:checkerboard_grid}
\end{figure*}

\begin{table*}[t]
\centering
\caption{Comparison of baseline method \citep{hendrycks2016baseline} and confidence-based thresholding. All models are trained on SVHN, which is used as the in-distribution dataset (average of 5 runs). All values are shown in percentages. $\downarrow$ indicates that lower values are better, while $\uparrow$ indicates that higher scores are better.}
\vspace{0.1in}
\begin{tabular}{llccccc}
\hline
 & \begin{tabular}[c]{@{}l@{}}\textbf{Out-of-distribution}\\ \textbf{dataset}\end{tabular} & \multicolumn{1}{c}{\begin{tabular}[c]{@{}c@{}}\textbf{FPR}\\ \textbf{(95\% TPR)}\\ $\downarrow$\end{tabular}} & \multicolumn{1}{c}{\begin{tabular}[c]{@{}c@{}}\textbf{Detection}\\ \textbf{Error}\\ $\downarrow$\end{tabular}} & \multicolumn{1}{c}{\begin{tabular}[c]{@{}c@{}}\textbf{AUROC}\\ \\ $\uparrow$\end{tabular}} & \multicolumn{1}{c}{\begin{tabular}[c]{@{}c@{}}\textbf{AUPR}\\ \textbf{In}\\ $\uparrow$\end{tabular}} & \multicolumn{1}{c}{\begin{tabular}[c]{@{}c@{}}\textbf{AUPR}\\ \textbf{Out}\\ $\uparrow$\end{tabular}} \\
\hline
& & \multicolumn{5}{c}{\textbf{Baseline \citep{hendrycks2016baseline}/Confidence Thresholding}} \\
\cline{3-7}
\multirow{7}{*}{\textbf{DenseNet-BC}} & TinyImageNet (crop) & 12.9/\textbf{3.1} & 7.0/\textbf{4.0} & 97.4/\textbf{99.2} & 99.0/\textbf{99.5} & 93.2/\textbf{98.0} \\
 & TinyImageNet (resize) & 7.2/\textbf{1.5} & 5.3/\textbf{2.8} & 98.4/\textbf{99.5} & 99.4/\textbf{99.8} & 95.6/\textbf{98.7} \\
 & LSUN (crop) & 28.3/\textbf{14.0} & 11.7/\textbf{8.4} & 93.7/\textbf{96.6} & 96.8/\textbf{98.1} & 85.6/\textbf{93.2} \\
 & LSUN (resize) & 6.0/\textbf{1.0} & 4.9/\textbf{2.3} & 98.6/\textbf{99.7} & 99.5/\textbf{99.9} & 96.0/\textbf{99.0} \\
 & iSUN & 6.0/\textbf{0.9} & 4.9/\textbf{2.3} & 98.6/\textbf{99.7} & 99.5/\textbf{99.9} & 95.7/\textbf{98.8} \\
 & Uniform & 11.9/\textbf{0.1} & 5.2/\textbf{1.2} & 97.9/\textbf{99.9} & 99.3/\textbf{100.0} & 93.4/\textbf{99.6} \\
 & Gaussian & 7.4/\textbf{0.0} & 4.1/\textbf{0.9} & 98.5/\textbf{99.9} & 99.5/\textbf{100.0} & 95.1/\textbf{99.7} \\
 & All Images & 12.2/\textbf{4.2} & 7.2/\textbf{4.5} & 97.3/\textbf{98.9} & 95.1/\textbf{97.4} & 98.4/\textbf{99.4} \\
\hline
{\multirow{7}{*}{\textbf{WRN-16-8}}} & TinyImageNet (crop) & 16.3/\textbf{4.1} & 7.8/\textbf{4.4} & 96.9/\textbf{99.1} & 98.7/\textbf{99.5} & 91.4/\textbf{98.1} \\
\multicolumn{1}{c}{} & TinyImageNet (resize) & 10.6/\textbf{1.5} & 6.1/\textbf{2.7} & 97.8/\textbf{99.6} & 99.2/\textbf{99.8} & 93.6/\textbf{99.2} \\
\multicolumn{1}{c}{} & LSUN (crop) & 31.9/\textbf{18.7} & 12.5/\textbf{10.5} & 93.0/\textbf{95.5} & 96.4/\textbf{97.5} & 84.0/\textbf{91.8} \\
\multicolumn{1}{c}{} & LSUN (resize) & 9.5/\textbf{0.6} & 5.8/\textbf{1.8} & 98.0/\textbf{99.8} & 99.3/\textbf{99.9} & 94.0/\textbf{99.5} \\
\multicolumn{1}{c}{} & iSUN & 9.6/\textbf{0.8} & 5.9/\textbf{2.1} & 98.0/\textbf{99.8} & 99.3/\textbf{99.9} & 93.4/\textbf{99.4} \\
\multicolumn{1}{c}{} & Uniform & 17.7/\textbf{0.3} & 7.1/\textbf{1.2} & 97.1/\textbf{99.9} & 99.0/\textbf{100.0} & 91.1/\textbf{99.6} \\
\multicolumn{1}{c}{} & Gaussian & 11.0/\textbf{0.2} & 5.8/\textbf{1.0} & 97.9/\textbf{99.9} & 99.3/\textbf{100.0} & 93.7/\textbf{99.8} \\
\multicolumn{1}{c}{} & All Images & 15.7/\textbf{5.3} & 7.9/\textbf{5.0} & 96.7/\textbf{98.7} & 94.1/\textbf{96.8} & 97.9/\textbf{99.4} \\
\hline
\multirow{7}{*}{\textbf{VGG13}} & TinyImageNet (crop) & 17.3/\textbf{4.3} & 7.7/\textbf{4.6} & 96.9/\textbf{99.2} & 98.8/\textbf{99.7} & 91.3/\textbf{98.1} \\
 & TinyImageNet (resize) & 11.4/\textbf{1.8} & 6.2/\textbf{3.1} & 97.8/\textbf{99.6} & 99.2/\textbf{99.8} & 93.7/\textbf{99.1} \\
 & LSUN (crop) & 22.7/\textbf{13.0} & 9.4/\textbf{7.8} & 95.6/\textbf{97.6} & 98.1/\textbf{99.0} & 88.6/\textbf{94.7} \\
 & LSUN (resize) & 9.4/\textbf{0.8} & 5.7/\textbf{2.0} & 98.1/\textbf{99.8} & 99.3/\textbf{99.9} & 94.3/\textbf{99.6} \\
 & iSUN & 10.0/\textbf{1.0} & 6.0/\textbf{2.2} & 98.0/\textbf{99.8} & 99.3/\textbf{99.9} & 93.7/\textbf{99.5} \\
 & Uniform & 20.0/\textbf{0.5} & 7.3/\textbf{1.4} & 96.8/\textbf{99.9} & 98.9/\textbf{100.0} & 90.2/\textbf{99.7} \\
 & Gaussian & 12.9/\textbf{0.3} & 6.0/\textbf{0.9} & 97.8/\textbf{99.9} & 99.2/\textbf{100.0} & 93.1/\textbf{99.9} \\
 & All Images & 14.2/\textbf{4.3} & 7.1/\textbf{4.6} & 97.3/\textbf{99.2} & 95.9/\textbf{98.5} & 98.2/\textbf{99.6} \\
\hline
\end{tabular}
\label{table:baseline_confidence_svhn}
\end{table*}

\begin{table*}[t]
\centering
\caption{Comparison of baseline method \citep{hendrycks2016baseline} and confidence-based thresholding. All models are trained on CIFAR-10, which is used as the in-distribution dataset (average of 5 runs). All values are shown in percentages. $\downarrow$ indicates that lower values are better, while $\uparrow$ indicates that higher scores are better.}
\vspace{0.1in}
\begin{tabular}{llccccc}
\hline
 & \begin{tabular}[c]{@{}l@{}}\textbf{Out-of-distribution}\\ \textbf{dataset}\end{tabular} & \multicolumn{1}{c}{\begin{tabular}[c]{@{}c@{}}\textbf{FPR}\\ \textbf{(95\% TPR)}\\ $\downarrow$\end{tabular}} & \multicolumn{1}{c}{\begin{tabular}[c]{@{}c@{}}\textbf{Detection}\\ \textbf{Error}\\ $\downarrow$\end{tabular}} & \multicolumn{1}{c}{\begin{tabular}[c]{@{}c@{}}\textbf{AUROC}\\ \\ $\uparrow$\end{tabular}} & \multicolumn{1}{c}{\begin{tabular}[c]{@{}c@{}}\textbf{AUPR}\\ \textbf{In}\\ $\uparrow$\end{tabular}} & \multicolumn{1}{c}{\begin{tabular}[c]{@{}c@{}}\textbf{AUPR}\\ \textbf{Out}\\ $\uparrow$\end{tabular}} \\
\hline
& & \multicolumn{5}{c}{\textbf{Baseline \citep{hendrycks2016baseline}/Confidence Thresholding}} \\
\cline{3-7}
\multirow{7}{*}{\textbf{DenseNet-BC}} & TinyImageNet (crop) & 42.6/\textbf{29.1} & 11.9/\textbf{11.0} & 93.8/\textbf{95.1} & 95.2/\textbf{95.8} & 91.8/\textbf{94.1} \\
 & TinyImageNet (resize) & 44.9/\textbf{33.8} & 12.8/\textbf{12.3} & 93.2/\textbf{94.2} & 94.6/\textbf{95.0} & 91.2/\textbf{93.0} \\
 & LSUN (crop) & 37.2/\textbf{19.6} & 10.8/\textbf{9.2} & 94.8/\textbf{96.6} & 95.9/\textbf{97.0} & 93.1/\textbf{96.1} \\
 & LSUN (resize) & 38.6/\textbf{30.7} & 10.8/\textbf{10.3} & 94.6/\textbf{95.4} & 95.9/\textbf{96.4} & 92.8/\textbf{93.9} \\
 & iSUN & 41.4/\textbf{31.6} & 11.6/\textbf{11.0} & 94.1/\textbf{95.0} & 95.8/\textbf{96.3} & 91.3/\textbf{93.0} \\
 & Uniform & 74.9/\textbf{11.7} & 18.2/\textbf{3.3} & 77.0/\textbf{97.7} & 82.8/\textbf{98.6} & 71.5/\textbf{94.3} \\
 & Gaussian & 66.1/\textbf{57.3} & 18.8/\textbf{8.5} & 75.0/\textbf{92.0} & 81.5/\textbf{95.1} & 71.5/\textbf{84.0} \\
 & All Images & 40.9/\textbf{28.9} & 11.6/\textbf{10.9} & 94.1/\textbf{95.3} & 87.6/\textbf{88.1} & 98.3/\textbf{98.7} \\
\hline
{\multirow{7}{*}{\textbf{WRN-28-10}}} & TinyImageNet (crop) & 36.7/\textbf{23.3} & 12.8/\textbf{10.0} & 92.5/\textbf{95.6} & 91.2/\textbf{95.7} & 91.8/\textbf{94.8} \\
\multicolumn{1}{c}{} & TinyImageNet (resize) & 41.0/\textbf{26.6} & 14.3/\textbf{11.6} & 91.0/\textbf{94.5} & 88.9/\textbf{94.1} & 90.5/\textbf{94.0} \\
\multicolumn{1}{c}{} & LSUN (crop) & 31.6/\textbf{17.6} & 10.6/\textbf{7.9} & 94.4/\textbf{96.9} & 94.2/\textbf{97.3} & 93.3/\textbf{96.2} \\
\multicolumn{1}{c}{} & LSUN (resize) & 34.7/\textbf{24.0} & 11.7/\textbf{9.1} & 93.7/\textbf{96.0} & 93.4/\textbf{96.6} & 92.7/\textbf{94.5} \\
\multicolumn{1}{c}{} & iSUN & 36.7/\textbf{24.9} & 12.6/\textbf{9.8} & 92.8/\textbf{95.7} & 92.6/\textbf{96.5} & 91.1/\textbf{94.0} \\
\multicolumn{1}{c}{} & Uniform & 60.4/\textbf{17.9} & 11.3/\textbf{3.8} & 91.3/\textbf{97.4} & 93.9/\textbf{98.4} & 84.2/\textbf{94.3} \\
\multicolumn{1}{c}{} & Gaussian & 68.6/\textbf{30.4} & 12.7/\textbf{4.9} & 89.8/\textbf{96.5} & 92.8/\textbf{97.8} & 81.6/\textbf{92.3} \\
\multicolumn{1}{c}{}  & All Images & 36.1/\textbf{23.3} & 12.4/\textbf{9.7} & 92.9/\textbf{95.7} & 73.3/\textbf{86.7} & 98.1/\textbf{98.8} \\
\hline
\multirow{7}{*}{\textbf{VGG13}} & TinyImageNet (crop) & 42.4/\textbf{20.8} & 11.7/\textbf{9.4} & 93.8/\textbf{96.8} & 94.9/\textbf{97.2} & 92.1/\textbf{96.4} \\
 & TinyImageNet (resize) & 43.8/\textbf{18.4} & 12.0/\textbf{9.4} & 93.5/\textbf{97.0} & 94.6/\textbf{97.3} & 91.7/\textbf{96.9} \\
 & LSUN (crop) & 38.9/\textbf{23.6} & 11.4/\textbf{9.4} & 94.1/\textbf{96.6} & 95.1/\textbf{97.1} & 92.6/\textbf{96.1} \\
 & LSUN (resize) & 41.9/\textbf{16.4} & 11.5/\textbf{8.3} & 94.0/\textbf{97.5} & 95.1/\textbf{97.8} & 92.2/\textbf{97.2} \\
 & iSUN & 41.2/\textbf{16.3} & 11.4/\textbf{8.5} & 94.0/\textbf{97.5} & 95.5/\textbf{98.0} & 91.5/\textbf{96.9} \\
 & Uniform & \textbf{20.7}/65.7 & \textbf{5.4}/7.8 & \textbf{97.0}/92.8 & \textbf{98.0}/95.7 & \textbf{94.8}/84.0 \\
 & Gaussian & \textbf{31.1}/84.1 & \textbf{6.2}/9.5 & \textbf{96.0}/90.1 & \textbf{97.4}/94.2 & \textbf{92.3}/78.9 \\
 & All Images & 41.6/\textbf{19.2} & 11.7/\textbf{9.1} & 93.9/\textbf{97.1} & 85.5/\textbf{92.0} & 98.2/\textbf{99.3} \\
\hline
\end{tabular}
\label{table:baseline_confidence_cifar10}
\end{table*}

\end{document}